%% file: main.tex
\documentclass[10pt,twocolumn,letterpaper]{article}

\usepackage[pagenumbers]{cvpr} 
\usepackage{caption}

\input{configs}

\newcommand{\toolA}[0]{\adjustbox{valign=c}{\includegraphics[scale=0.03]{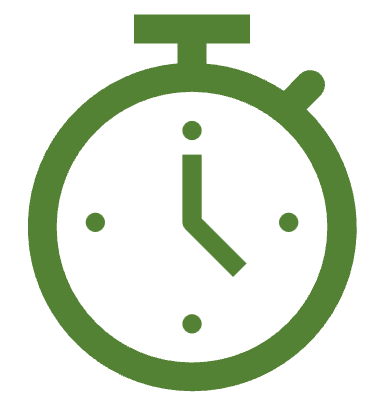}}}
\newcommand{\toolB}[0]{\adjustbox{valign=c}{\includegraphics[scale=0.03]{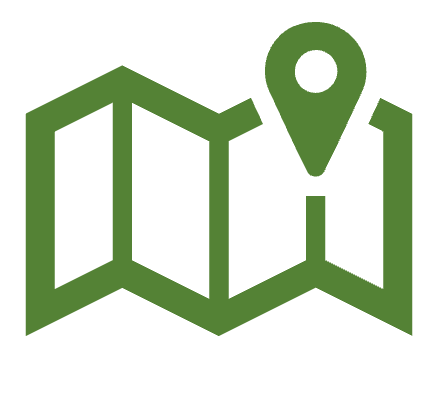}}}
\newcommand{\toolC}[0]{\adjustbox{valign=c}{\includegraphics[scale=0.03]{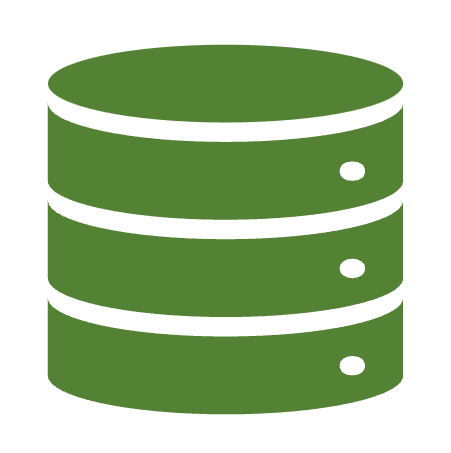}}}
\newcommand{\toolD}[0]{\adjustbox{valign=c}{\includegraphics[scale=0.03]{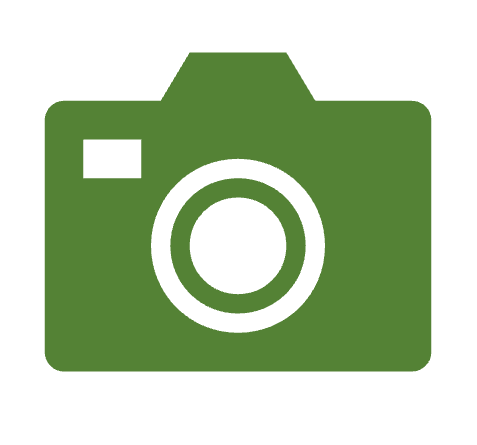}}}
\newcommand{\toolchat}[0]{\adjustbox{valign=c}{\includegraphics[scale=0.05]{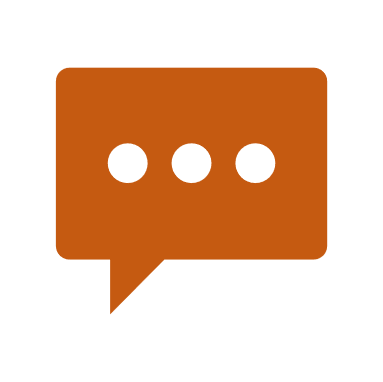}}}
\newcommand{\toolclose}[0]{\adjustbox{valign=c}{\includegraphics[scale=0.05]{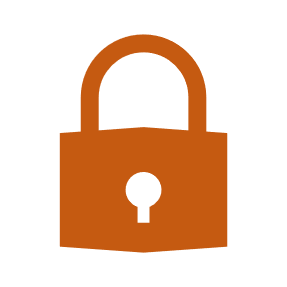}}}
\newcommand{\toolopen}[0]{\adjustbox{valign=c}{\includegraphics[scale=0.05]{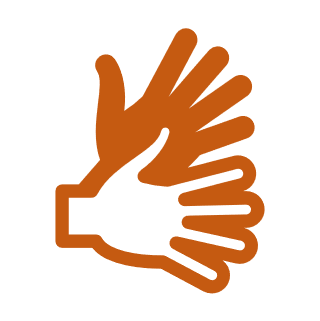}}}
\newcommand{\toolgoto}[0]{\adjustbox{valign=c}{\includegraphics[scale=0.05]{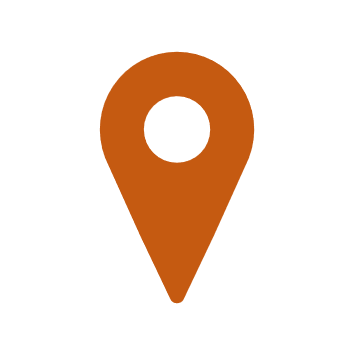}}}
\newcommand{\toolpick}[0]{\adjustbox{valign=c}{\includegraphics[scale=0.05]{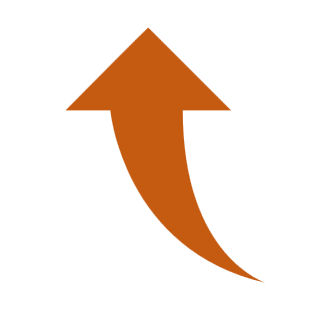}}}
\newcommand{\toolplace}[0]{\adjustbox{valign=c}{\includegraphics[scale=0.05]{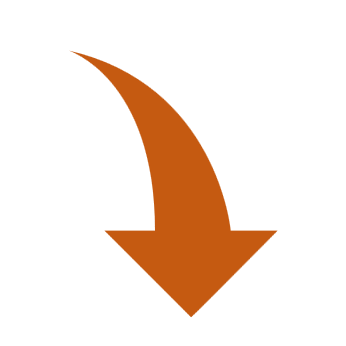}}}
\newcommand{\toolsearch}[0]{\adjustbox{valign=c}{\includegraphics[scale=0.05]{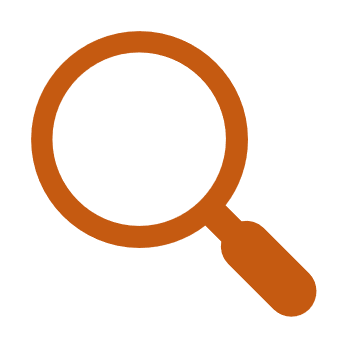}}}
\newcommand{\embvideoagent}[0]{\adjustbox{valign=c}{\includegraphics[scale=0.05]{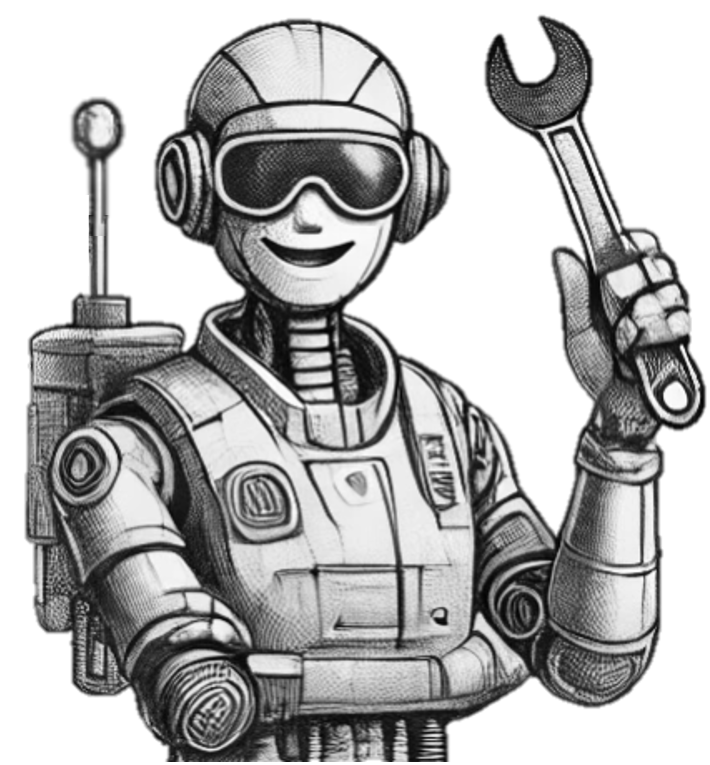}}}

\definecolor{cvprblue}{rgb}{0.21,0.49,0.74}
\usepackage[pagebackref,breaklinks,colorlinks,allcolors=cvprblue]{hyperref}

\def\paperID{6011} 
\def\confName{CVPR}
\def\confYear{2025}

\title{\embvideoagent\xspace Embodied VideoAgent: Persistent Memory from Egocentric \\ Videos and Embodied Sensors Enables Dynamic Scene Understanding}

\author{
Yue Fan\textsuperscript{\rm 1$^\star$},
Xiaojian Ma\textsuperscript{\rm 1$^\star$$^\dag$},
Rongpeng Su\textsuperscript{\rm 1,2},
Jun Guo\textsuperscript{\rm 1,3},
Rujie Wu\textsuperscript{\rm 1,4},
Xi Chen\textsuperscript{\rm 1},
Qing Li\textsuperscript{\rm 1$^\dag$} \\
 \small \textsuperscript{\rm 1}State Key Laboratory of General Artificial Intelligence, BIGAI, Beijing, China \\
 \small \textsuperscript{\rm 2}University of Science and Technology of China 
 \small \textsuperscript{\rm 3}Tsinghua University 
 \small \textsuperscript{\rm 4}Peking University \\
 \small \{fanyue,maxiaojian,liqing\}@bigai.ai\\
 \small \href{https://embodied-videoagent.github.io/}{https://embodied-videoagent.github.io}
}

\begin{document}

\maketitle
\renewcommand{\thefootnote}{}
\footnotetext{$^\star$Equal contribution.}
\footnotetext{$^\dag$Corresponding authors.}
\renewcommand{\thefootnote}{\arabic{footnote}}
\setcounter{footnote}{0}


\begin{abstract}
This paper investigates the problem of understanding dynamic 3D scenes from egocentric observations, a key challenge in robotics and embodied AI. Unlike prior studies that explored this as long-form video understanding and utilized egocentric video only, we instead propose an LLM-based agent, \method, which constructs scene memory from both egocentric video and embodied sensory inputs (\eg depth and pose sensing). We further introduce a VLM-based approach to automatically update the memory when actions or activities over objects are perceived. \method attains significant advantages over counterparts in challenging reasoning and planning tasks in 3D scenes, achieving gains of 4.9\% on Ego4D-VQ3D, 5.8\% on OpenEQA, and 11.7\% on EnvQA. We have also demonstrated its potential in various embodied AI tasks including generating embodied interactions and perception for robot manipulation. The code and demo will be made public.

\end{abstract}

\section{Introduction}

Understanding dynamic 3D scenes is crucial to the development of generally capable embodied AI~\cite{huang2023embodied,wang2023jarvis,wang2024omnijarvis,wang2023describe}. In this paper, we investigate approaching this problem using egocentric observations~\cite{mangalam2024egoschema,ego4d,jia2022egotaskqa,videoagent,Damen2022RESCALING,pramanick2023egovlpv2,lin2022egocentric}, which is one of the most intuitive way of how humans and robots perceive the world around them. The key challenges include: 1) Making sense of environments from lengthy egocentric videos and other forms of embodied sensory inputs (depth maps, camera poses, \etc)~\cite{huang2023embodied,sajjadi2022scene,sitzmann2019scene}; 2) Handling dynamic environments as actions and activities might be performed by embodied agents themselves and other co-habited characters~\cite{ego4d,mangalam2024egoschema,Damen2022RESCALING}; 3) Maintaining a persistent memory about the scene that allows frequent update over time~\cite{gu2024conceptgraphs,jatavallabhula2023conceptfusion,rivera2024conceptagent}. However, existing efforts on this front mostly adopt end-to-end pretrained multimodal large models (MLMs)~\cite{videollava,llamavid,jxma_vlm_multimodal_2023,zhu20233d,huang2023embodied,zhu2025unifying,liu2023visual}. Their capabilities of handling long-form videos and embodied sensory observations have been questioned by several prior studies~\cite{tapaswi2016movieqa,miech2019howto100m,wu2021towards,korbar2023text}, especially when the scene being depicted is highly volatile with complex events and spatial-temporal dependencies~\cite{han2022temporal,jia2022egotaskqa,jia2020lemma}. Some sophisticated MLMs coming out recently have attained great progress in understanding long-form videos and the underlying embodied scenes~\cite{team2023gemini,wiles2022compressed,team2024gemini}, but the computation cost can grow at a prohibitively expensive rate. All these issues have hindered the progress of deploying robust dynamic scene perception pipelines on edge devices like robots in the wild.

\begin{figure}[t!]
    \centering
    \includegraphics[width=\linewidth]{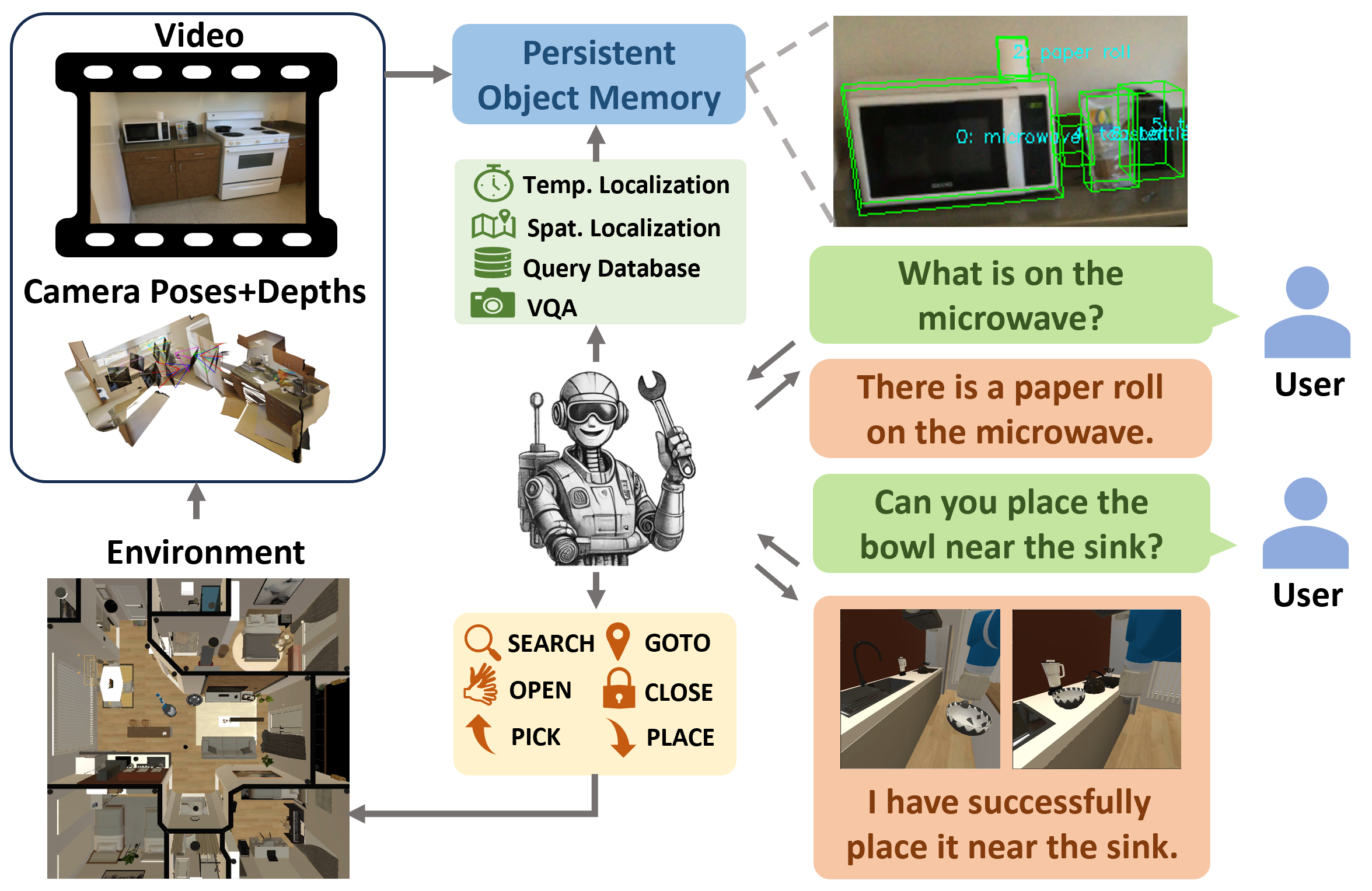}
    \caption{\method is a multimodal agent that 1) builds scene memory from both egocentric video and embodied sensory input; 2) utilizes multiple tools 
    to query this memory; 3) activates embodied action primitives to interact with the environments, effectively fulfills various user requests.}
    \label{fig:teaser}
    \vskip -0.2in
\end{figure}

Unlike these end-to-end models, there has been rapid development in another family of multimodal understanding approaches -- \textit{multimodal agents}~\cite{gao2023clova,videoagent,suris2023vipergpt,wu2023visual,gupta2023visual}. These methods utilize the advanced reasoning and \textit{tool-usage} ability of pretrained large language models (LLMs) and solve intricate multimodal tasks by calling several tool models (object understanding, question answering, \etc) interactively, alleviating the issue of expensive training and inference cost. Recently, they have been ported to long-form video understanding and have demonstrated remarkable performances and cost-efficiency over end-to-end counterparts~\cite{videoagent,suris2023vipergpt,gao2023clova}. Their key idea is to construct a temporal memory from the video and invoke several tools to query the memory. However, extending them to understanding dynamic 3D scenes is \textbf{non-trivial}. 
The challenges as mentioned earlier require the system to have a comprehensive yet precise understanding of objects in the scene subjected to constant change due to various actions and activities being performed by the embodied agents and other characters. Our early explorations (see \autoref{sec:experiments}) on simply applying these systems (\eg VideoAgent~\cite{videoagent}) to dynamic 3D scenes in embodied environments have suggested that merely constructing scene memory from video with hand-crafted pipelines cannot meet the aforementioned requirements, especially on the precise understanding of objects and the support of dynamic memory update, leading to unsatisfactory performances on these tasks.

To this end, we propose \method, a simple yet effective multimodal agent for understanding dynamic 3D scenes. Our agent is based upon VideoAgent~\cite{videoagent}, a recent multimodal agent that can solve various video understanding tasks by constructing memory on long-form videos and performing LLM-based queries over the memory. Our \textbf{key innovation} is to augment it with two novel designs for dynamic scenes in embodied environments: 1) a \textbf{persistent object memory} that is constructed from \textit{both} egocentric video and embodied sensory input (depth maps and camera poses). Fusing video with these modalities could help build more precise memory on scene objects, which is crucial to embodied reasoning and planning; 2) a \textbf{VLM-based memory update mechanism} that automatically identifies relevant objects and their state changes when actions and activities are being perceived, then promptly updates the corresponding entries in the persistent object memory. In addition to understanding tasks, we explore the potential of \method in generating embodied user-assistant interactions. Specifically, we introduce an LLM-based multi-agent framework~\cite{jxma_llm_vla_vlm_mas_multiagent_2023}, where a user agent proposes tasks, and an assistant agent (effectively an \method) progressively explores the scene to complete these tasks. The assistant provides feedback to the user while tracking its evolving understanding of the dynamic scene using persistent object memory.

\begin{figure*}[t!]
    \centering
    \includegraphics[width=\linewidth]{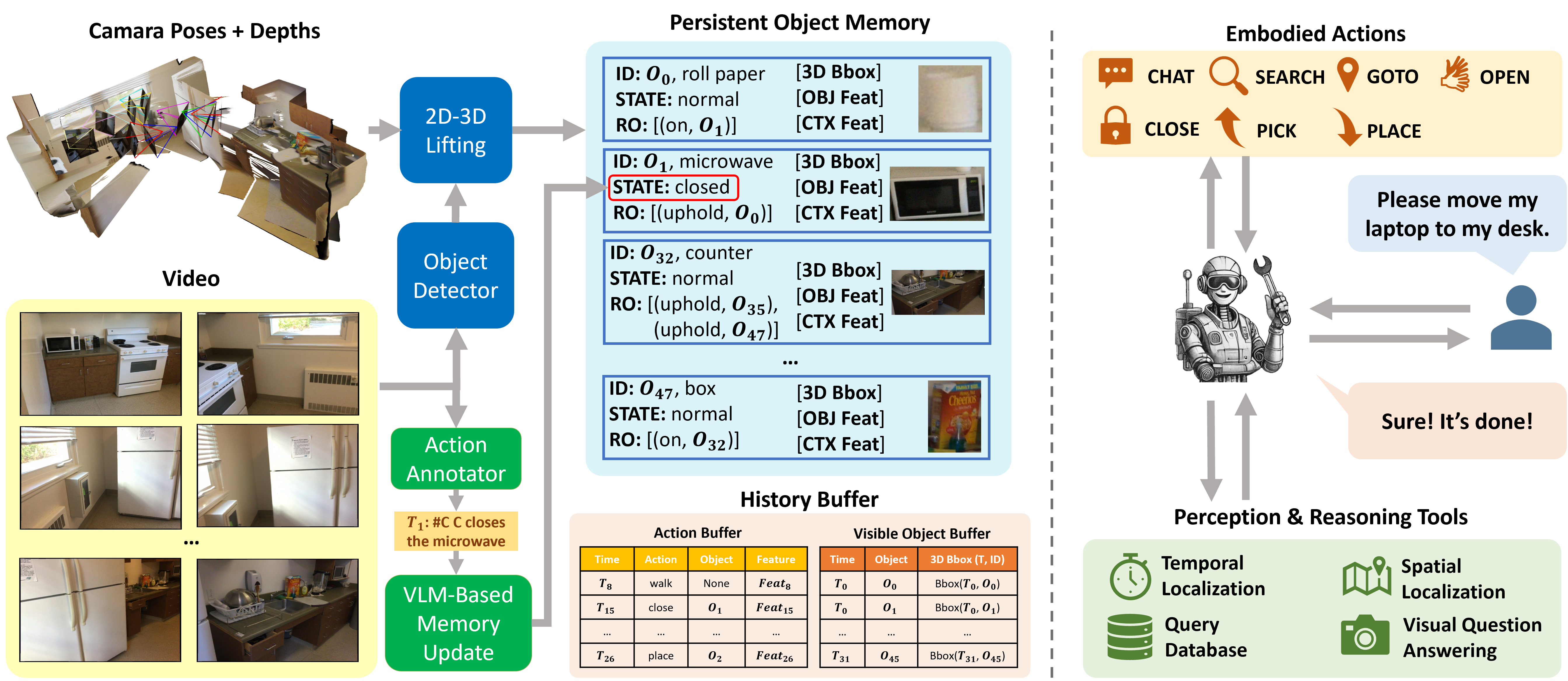}
    \caption{An overview of \method. Left: We first translate the egocentric video and embodied sensory input (depth maps and camera poses) into structured representations: \textit{persistent object memory} and history buffer. While the memory can be updated using VLM to support dynamic scenes where actions are being performed constantly; Right: the LLM within \method is prompted to fulfill the user's request by interactively invoking tools to query the memory and calling embodied action primitives to complete the task.}
    \label{fig:method}
    \vskip -0.2in
\end{figure*}

We conduct extensive evaluations of \method on several embodied scene understanding tasks, including object localization from free-form queries in egocentric views of dynamic scenes with Ego4D-VQ3D~\cite{ego4d,egoloc}, general question answering on embodied 3D scenes with OpenEQA~\cite{openeqa}, and question answering on long-form embodied robot-environment interactions with EnvQA~\cite{envqa}. We compare \method against both the canonical end-to-end multimodal LLMs and other multimodal agents. Results confirm the advantages of \method: achieving gains of 4.9\% on Ego4D-VQ3D, 5.8\% on OpenEQA, and 11.7\% on EnvQA. Our further exploration has covered its applications in generating embodied interaction and perception for robot manipulation.

Our contributions can be summarized as follows:
\setlength{\leftmargini}{0.85em}
\begin{itemize}[topsep=0pt]
\item We propose a persistent object memory along with a VLM-based automatic memory update method to construct and maintain comprehensive yet precise memory of dynamic 3D scenes from both egocentric and embodied (depth maps, camera poses) sensory observations. 
\item With the resulting agent, \method, we further develop an LLM-based multi-agent framework that can produce embodied user-assistant interactions, where the user proposes tasks and the assistant (an \method) progressively explores the scene to complete these tasks.
\item We conduct thorough evaluations of \method on various embodied scene understanding tasks against both end-to-end multimodal LLMs and multimodal agent baselines, siding with applications in two embodied AI tasks, demonstrating the effectiveness of \method.
\end{itemize}


\section{Embodied VideoAgent}\label{sec:method}


We illustrate the proposed \method in \autoref{fig:method}. Since our agent is based upon VideoAgent~\cite{videoagent}, we will first quickly recap its key ideas (\autoref{sec:videoagent}), then move on to cover the key new memory and tool design, including persistent object memory and VLM-based memory update method (\autoref{sec:object_mem}). Finally, we will detail how we utilize \method for generating embodied user-assistant interactions (\autoref{sec:two_agent}).

\subsection{Recap: VideoAgent}\label{sec:videoagent}

VideoAgent~\cite{videoagent} adopts the following pipeline: given a video $V$ sliced into $n$ segments $[v_1, \dots, v_n]$, it first constructs a temporal memory $\mathcal{M}_T$, which captures the textual descriptions (and features) of each segment; and an object memory $\mathcal{M}_O$,  which tracks and store the occurrences of objects and persons in the video. Then for any incoming task, an LLM decomposes it into several subtasks and invokes tool models to query the temporal and object memory. Finally, the responses of all tool-calling will be aggregated and sent to an LLM for a final answer. 

\noindent
\textbf{Temporal and Object Memory.~~} For temporal memory, it is effectively a table with $n$ rows, where $n$ is the number of short (2s) video segments. Each row has four columns: segment ID, caption of this segment $s_{\text{caption}}$, visual feature of this segment $e_{\text{video}}$ and text embedding of the caption $e_{\text{caption}}$. 
For object memory, it includes a SQL database and a feature table of all identified objects in the video. The SQL database has three columns: unique object ID, object category, and the segment IDs where the object occurs. Its construction requires object detection, tracking, and re-ID. The feature table stores the CLIP feature of the object image.

\noindent
\textbf{Tool-usage and Inference.~~} VideoAgent utilizes four tools -- caption retrieval, segment localization, visual question answering, and object memory query to access the temporal and object memory. The inference process is straightforward: given an input query, VideoAgent selects an appropriate tool, invokes it, and stores the result in a buffer. This loop continues until VideoAgent either decides to stop or reaches a predefined maximum number of steps, after which it generates a final response based on the buffer’s content.

Readers are encouraged to refer to~\cite{videoagent} for more details.

\subsection{Memory and Tools of Embodied VideoAgent}\label{sec:object_mem}


\begin{figure*}[t!]
    \centering
    \includegraphics[width=\textwidth]{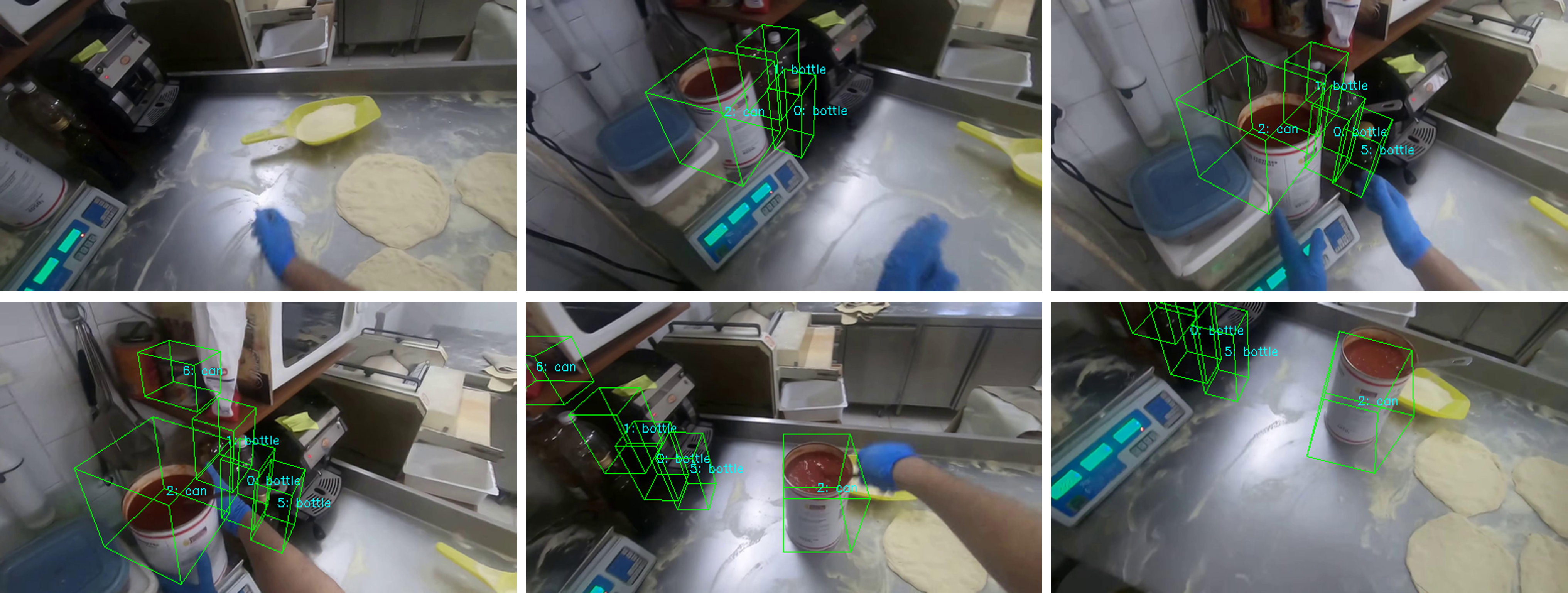}
    \caption{Visualization of the entries in \textit{persistent object memory} $\mathcal{M}_O$. Each 3D bounding box corresponds to an entry in the memory. As the video proceeds, objects (\eg the large canned tomato paste) can be tracked/re-IDed and have their memory entries updated.}
    \label{fig:object_mem_viz}
    \vskip -0.2in
\end{figure*}

\method adopts the following memory and tool design upon its predecessor VideoAgent: given an egocentric video downsampled to $n$ frames $V = [I_1, \dots, I_n]$, with depth map and camera 6D pose of each frame $D = [(d_1, p_1), \dots, (d_n, p_n)]$ ($d$ and $p$ denote depth map and camera pose, respectively), it constructs the original temporal memory $\mathcal{M}_T$ of VideoAgent (not shown in \autoref{fig:method}), the newly introduced persistent object memory $\mathcal{M}_O$, and two simple history buffers. Four tools (\toolC\xspace\texttt{query\_db}, \toolA\xspace\texttt{temporal\_loc}, \toolB\xspace\texttt{spatial\_loc}, \toolD\xspace\texttt{vqa}) can be invoked to access the memory. Several embodied action primitives are available to be called to interact with the physical environment. Details can be found below:

\noindent
\textbf{Persistent Object Memory $\mathcal{M}_O$.~~} It maintains an entry for each perceived object in the 3D scene. Each object entry includes the following fields: a unique object identifier $O_i$ with object category (\textbf{ID}), a state description of the object (\textbf{STATE}), a list of \underline{\textbf{r}}elated \underline{\textbf{o}}bjects and their relations (\textbf{RO}), 3D bounding box of the object(\textbf{3D Bbox}), visual feature of the object (\textbf{OBJ Feat}) and visual feature of the environment context where the object locates(\textbf{CTX Feat}). These fields provide comprehensive details of scene objects and their surroundings.

\noindent
\textbf{Construction of $\mathcal{M}_O$.~~} Given an incoming 2D egocentric frame $I_i$, depth map $d_i$ and camera pose $p_i$, we first use an open-vocabulary object detection model called YOLO-world~\cite{yoloworld} to extract objects and their categories (the \textbf{ID} field) from 2D frame $I_i$. Its state description will be initialized as ``normal'' (the \textbf{STATE} field). The CLIP feature of frame $I_i$ and the cropped object picture using the 2D bounding box will become the \textbf{CTX Feat} field and the \textbf{OBJ Feat} field, respectively. Further, by utilizing the depth map $d_i$ along with camera pose $p_i$, we can obtain the object's 3D bounding box using 2D-3D lifting (projection)~\cite{guo2024semantic} (the \textbf{3D Bbox} field). Then we follow the prior practice~\cite{jia2025sceneverse,zhu20233d,huang2023embodied} to extract the relations among the detected objects using their 3D bounding boxes (the \textbf{RO} field). So far, only two pairs of relations ``on/uphold'' and ``in/contain'' are considered. To avoid duplicated object entries in $\mathcal{M}_O$, an object re-ID procedure~\cite{videoagent} is also conducted before inserting an object as a new entry. Objects are considered identical based on their proximity in both visual appearance and 3D location. Once an object is re-IDed to an existing object entry, we will update its \textbf{3D Bbox}, \textbf{Obj Feat}, and \textbf{CTX Feat} fields using moving average, while the \textbf{RO} field will be re-computed along with other objects detected in the current frame. Due to space constraints, details of our re-ID algorithm and object entry update after re-ID are provided in \supp. A visualization of how object entries are created and updated in $\mathcal{M}_O$ can be found in \autoref{fig:object_mem_viz}.

\begin{figure}[t!]
    \centering
    \includegraphics[width=\columnwidth]{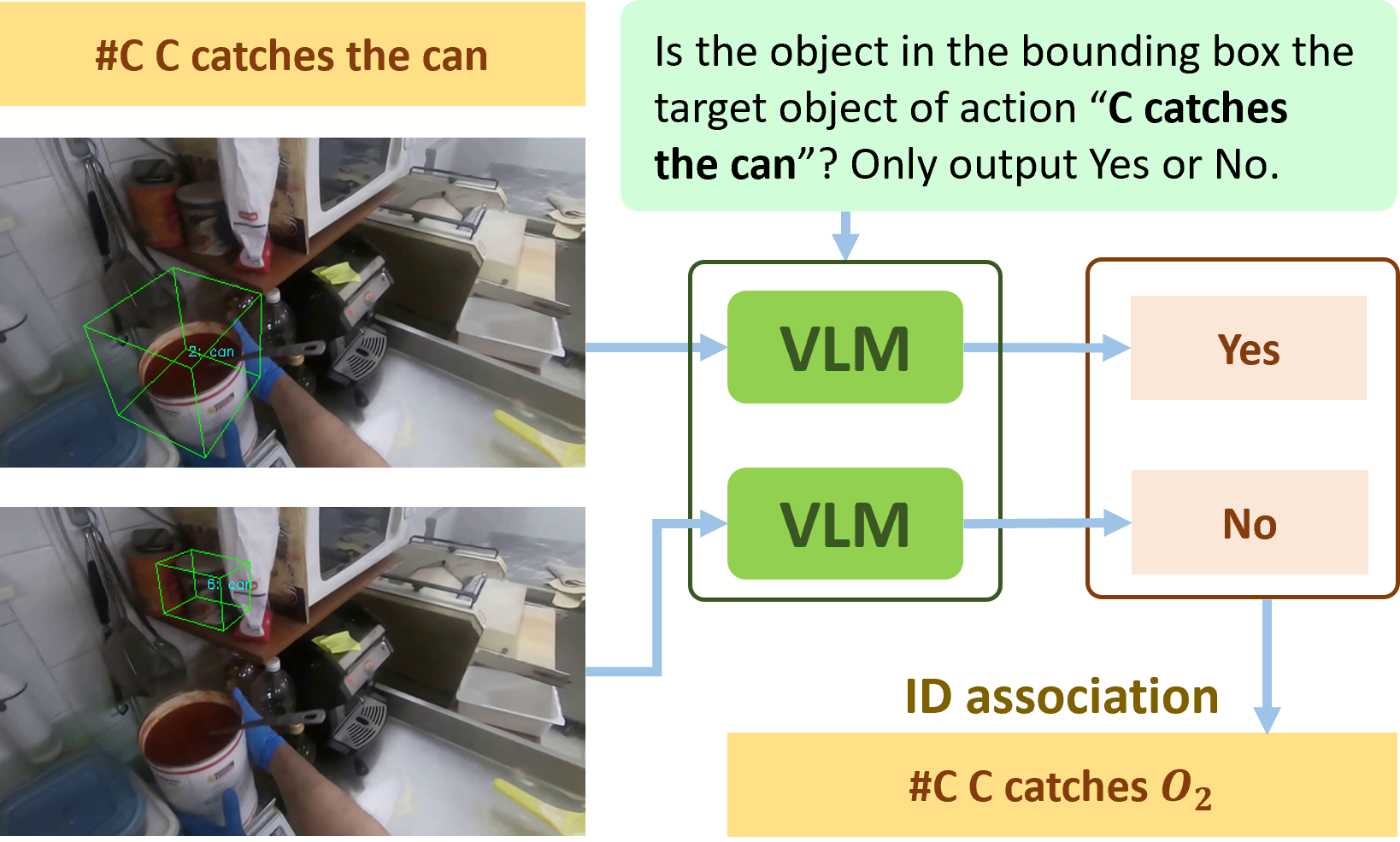}
    \caption{An illustration of our VLM-based memory update method. This approach effectively prompts the VLM to associate an action with relevant object entries in memory through visual prompting, identifying the entries corresponding to the action’s target objects.}
    \label{fig:vlm_update}
    \vskip -0.2in
\end{figure}

\noindent
\textbf{Memory Update with VLM.~~} A key challenge in persistent object memory lies in updating memory when actions are performed on objects, especially under conditions of visual occlusion (e.g., hand-can interaction in \autoref{fig:object_mem_viz}). We address this issue by leveraging action information and vision-language models (VLMs). As shown in \autoref{fig:vlm_update}, when an action occurs (e.g., ``C catches the can''), we first retrieve relevant object entries in $\mathcal{M}_O$ associated with the ``can'' that are visible in the current frame (in this example, two entries). For each entry, we render its 3D bounding box onto the frame and prompt the VLM to determine if the object within the box is the action’s target. Such visual prompting~\cite{yang2023setofmark,cai2024rocket} associates the action with corresponding entries in the object memory. Finally, we programmatically update these entries, such as modifying the \textbf{STATE} field to ``in-hand'' since the action is ``catches the can''. Additional details on the programmatic update are provided in \supp.

\noindent
\textbf{History Buffer.~~} In addition to persistent object memory that provides real-time information on current scene objects, we found that maintaining a simple record of past perception and action history further enhances dynamic scene understanding. For this purpose, we introduce two history buffers: an \textbf{action buffer}, which logs each action performed along with the action timestamp, action name, target object ID (identified using the VLM-based method), and the CLIP feature of the current frame; and a \textbf{visible object buffer}, which logs each detected object along with the detection timestamp, object ID, and 3D bounding box. These buffers are also referenced by the tools described later.

\noindent
\textbf{Tools and Embodied Action Primitives.~~} We equip \method with four tools: \toolC\xspace\texttt{query\_db($\cdot$)}, which processes natural language queries to retrieve the top-10 matching object entries by searching both the persistent object memory and history buffers; \toolA\xspace\texttt{temporal\_loc($\cdot$)}, inherited from VideoAgent, which maps natural language queries to specific video timesteps; \toolB\xspace\texttt{spatial\_loc($\cdot$)}, which provides a 3D scene location (aligned with the camera’s coordinate system) based on object and room queries; and \toolD\xspace\texttt{vqa($\cdot$)}, which answers open-ended questions about a given frame. Additionally, the agent can perform seven embodied action primitives: \texttt{chat()} for user interaction; \texttt{search($\cdot$)} to conduct exhaustive scene searches for specified objects; \texttt{goto($\cdot$)} for location navigation; and \texttt{open($\cdot$)}, \texttt{close($\cdot$)}, \texttt{pick($\cdot$)}, and \texttt{place($\cdot$)} for object interactions. Further implementation details on tools and action primitives are provided in \supp.

\noindent
\textbf{Note on camera poses.~~}While readers may view the requirement for precise 6D camera poses as idealized for real-world embodied agent settings, \method demonstrates robustness to pose estimation noise. In our experiments, camera poses for Ego4D-VQ3D (\autoref{sec:exp_ego4d}) and EnvQA (\autoref{sec:exp_envqa}) are estimated using COLMAP~\cite{schoenberger2016sfm} and DUSt3R~\cite{wang2024dust3r}, respectively, which are inherently noisier than the ground truth poses available in OpenEQA (\autoref{sec:exp_openeqa}). Despite these variances, our agent consistently achieves substantial improvements over baselines (e.g., VideoAgent) across all three settings. In \supp, we also present additional results on OpenEQA using estimated noisy poses to further substantiate this robustness. We hypothesize that our memory and tool design provide redundancy, enabling tasks to be completed via multiple pathways, effectively bypassing potentially flawed memory entries or tools. More inference examples are provided in \supp.

\subsection{A Two-Agent Framework for Generating Embodied Interactions}\label{sec:two_agent}

\begin{figure}[t!]
    \centering
    \includegraphics[width=\columnwidth]{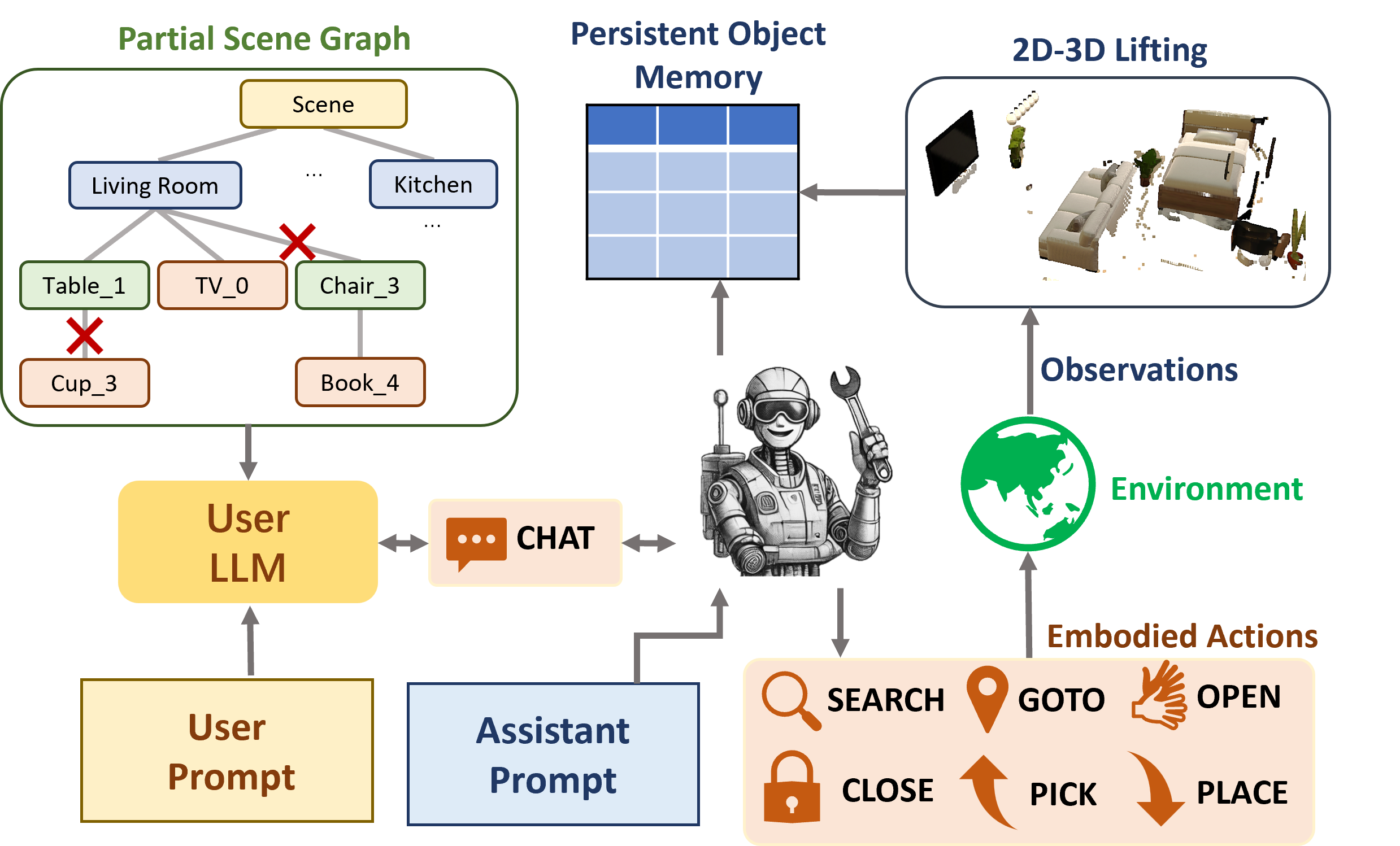}
    \caption{An overview of our synthetic embodied data collection framework. An LLM plays the user role and is prompted to propose engaging tasks based on a partial scene graph and the user's feedback, while the user, effectively a \method, explores the scene and fulfills the user's requests.}
    \label{fig:two_agent}
    \vskip -0.2in
\end{figure}

Collecting synthetic data for training foundation models, particularly embodied foundation models, has recently gained considerable interest~\cite{liu2024best}. We explore a novel approach with \method to gather synthetic embodied user-assistant interaction data. This dataset comprises episodes where a user interacts with an assistant within embodied environments. Drawing inspiration from prior multi-LLM-agent research~\cite{jxma_llm_vla_vlm_mas_multiagent_2023}, we use one LLM to emulate the user's role, while \method assumes the assistant's role, exploring the environment and fulfilling the user's diverse requests. An overview of this framework is shown in \autoref{fig:two_agent}. The user is prompted to propose varied and engaging tasks based on its limited scene graph knowledge—achieved by randomly trimming the full scene graph to stimulate curiosity—and the assistant’s feedback. Detailed prompting strategies are provided in \supp.

\section{Capabilities and Analysis}\label{sec:experiments}

We evaluate \method on various dynamic and embodied scene understanding tasks, including 3D object localization in dynamic scenes using Ego4D-VQ3D (\autoref{sec:exp_ego4d}), embodied question answering with OpenEQA (\autoref{sec:exp_openeqa}), and general question answering over embodied interactions on EnvQA (\autoref{sec:exp_envqa}). The performances are compared against state-of-the-art multimodal LLMs and multimodal agents. In \autoref{sec:exp_qualitative}, we demonstrate its application to two embodied AI tasks: generating embodied interactions and perception for robot manipulation.

\subsection{3D Object Localization}\label{sec:exp_ego4d}
We test \method on Ego4D Visual Queries 3D localization (VQ3D)\cite{ego4d}. Given an egocentric video depicting how a human subject interacts with a dynamic environment, an image of a target object, and a query frame, the task of VQ3D is to output the position of the target object at the time stamp of the query frame.
\subsubsection{Settings}

\noindent
\textbf{Baselines.~~}Two types of \method are tested: 1) retrieving the object with the highest visual score with the target image from the up-to-date object memory, denoted as \method(image); 2) retrieving the object with the same category as the target object, referred as \method(text). EgoLoc\cite{egoloc}, Ego4D*\cite{egoloc}, \method(text), and \method(image) are all based on the same precomputed camera poses and depth images provided by EgoLoc\cite{egoloc}, the 1st place on VQ3D challenge. Ego4D\cite{ego4d} denotes the baseline method in the benchmark paper. 

\noindent
\textbf{Metrics.~~}Among the metrics, Succ\% is the most important one that evaluates the success rate on all queries. Succ* (success rate on the answered queries only) and L2 (the average distance error) are all computed on the queries where the target object is detected by the method. The proportion of the answered queries to all queries is denoted as QwP\%.

\subsubsection{Results on VQ3D}
\autoref{ego4d_res} shows the results on VQ3D validation set. Overall, \method(image) achieves the highest success rate, surpassing EgoLoc by 5\%. We made the following observations:

\noindent
\textbf{Open-vocabulary object detector provides more candidate objects.~~}The higher QwP\% rate of \method(image): 92.07\% compared to that of EgoLoc, indicating the strong and robust performance of open-vocabulary object detection of \method empowered by YoloWorld\cite{yoloworld}. The better Succ*\% and L2 of EgoLoc can be attributed to its high-confidence predictions since these two metrics only evaluate the predicted queries. By contrast, \method(image) sacrifices little Succ*\% and L2 for more aggressive predictions on hard open-vocabulary queries, which finally results in the best QwP\% and Succ\%.

\noindent
\textbf{Visual Similarity is crucial for Object re-ID in a dynamic scene.~~}By only considering text for object retrieval, \method(text) has decent performance on VQ3D compared to Ego4D baseline, though not being competitive to \method(image). This can be attributed to the in-door settings of Ego4D videos, where functional objects (scissors, screwdrivers, etc) are usually clustered within the distance error of a successful detection. Applying visual-based object re-ID on object candidates boosts the performance, indicated by the large margin between \method(image) and \method(text), illustrating the effectiveness of the visual similarity score for object re-ID in dynamic scenes.

\begin{table}[]
\caption{Results of 3D object localization within dynamic scenes on the validation set of Ego4D-VQ3D\cite{ego4d}.}
\resizebox{\columnwidth}{!}{
\begin{tabular}{@{}cccccc@{}}
\toprule
\multicolumn{5}{c}{\textbf{Ego4D VQ3D}}                          
\\ \midrule
\multicolumn{1}{c|}{\textbf{Method}}              & \multicolumn{1}{l}{\textbf{Succ}$\%\uparrow$} & \multicolumn{1}{l}{\textbf{Succ*}$\%\uparrow$} & \multicolumn{1}{l}{\textbf{L2}$\downarrow$} &  \multicolumn{1}{l}{\textbf{QwP}$\%\uparrow$} \\ \midrule
\multicolumn{1}{c|}{EgoLoc}                       & \textbf{80.49}                       & \textbf{98.14}                        & \textbf{1.45}                     & \textbf{82.32}                         \\
\multicolumn{1}{c|}{Ego4D*}                       & 73.78                                & 91.45                                 & 2.05                                               & 80.49                                  \\
\multicolumn{1}{c|}{Ego4D}                        & 1.22                                 & 30.77                                 & 5.98                                                          & 1.83                                   \\
\multicolumn{1}{c|}{\textbf{E-VideoAgent(text)}}  & 53.05                                & \textbf{94.57}                        & 2.00                                                           & 56.10                                  \\
\multicolumn{1}{c|}{\textbf{E-VideoAgent(image)}} & \textbf{85.37}                       & 92.72                                 & \textbf{1.86}                                                & \textbf{92.07}                         \\ \bottomrule
\end{tabular}}
\label{ego4d_res}
\vskip -0.2in
\end{table}

\subsection{Embodied Question Answering}
Given an embodied episode in a scene, Embodied Question Answering requires the model to answer the question about the scene and embodied activities, such as ``what is the orange thing on the shelf to the right'', ``where did I leave my remote controller'', etc. \method is tested on OpenEQA\cite{openeqa} and EnvQA~\cite{envqa}, two recent benchmarks on open-ended embodied question answering. 
\subsubsection{Settings}
\noindent
\textbf{Baselines.~~} We equip \method with the four perception tools mentioned in \autoref{sec:object_mem}. For \toolD\xspace\texttt{vqa} tool, we tested InternVL2-8B\cite{internvl, internvl2} and GPT-4o, denoted as \method(InternVL2-8B) and \method(GPT-4o) respectively. Please note we compare with \textbf{zero-shot} baselines only following prior practices. On OpenEQA\cite{openeqa}, the baseline methods include 1) Large Video Language Models: Video-LLaVA\cite{videollava} and LLaMA-VID\cite{llamavid}; 2) multi-modal Agents: VideoAgent\cite{videoagent}, GPT-4 w/LLaVA-1.5\cite{openeqa} (which leverages frame captions) and GPT-4 w/CG\cite{openeqa} (which uses scene graph information). \method is tested on a subset of the original dataset due to cost issues, with the subset size being one-fifth of the original dataset. The questions in the subset are randomly selected. On EnvQA\cite{envqa}, we tested \method with Video-LLaVA\cite{videollava}, LLaMA-VID\cite{llamavid} and VideoAgent\cite{videoagent}. We tested these methods on three types of questions of EnvQA: States (e.g. "Where was the book moved?"), Events (e.g. ``what happened, after throwing soap bar and before throwing soap bar to hit shower door?''), and Orders (e.g. ``filling pot with water or use up soap bottle, which happened first''), with each type containing 200 questions.

\subsubsection{Results on OpenEQA}\label{sec:exp_openeqa}
\begin{table}[]
\caption{Results of embodied question answering on the EM-EQA split of OpenEQA\cite{openeqa}. Some scores are borrowed from the original benchmark paper.}
\resizebox{\columnwidth}{!}{
\begin{tabular}{@{}cccc@{}}
\toprule
\multicolumn{4}{c}{\textbf{OpenEQA}}                                                                        \\ \midrule
\multicolumn{1}{c|}{\textbf{Method}}                  & \textbf{ScanNet} & \textbf{HM3D}  & \textbf{ALL}   \\
\midrule
\multicolumn{1}{c|}{GPT-4 w/ LLaVA-1.5}               & \textbf{45.4}   & \textbf{40.0} & \textbf{43.6} \\
\multicolumn{1}{c|}{GPT-4 w/ CG}                      & 37.8            & 34.0          & 36.5          \\
\multicolumn{1}{c|}{Video-LLaVA}                       & 41.5            & 34.6          & 39.2          \\ 
\multicolumn{1}{c|}{LLaMA-VID}                       & 33.4            & 34.0          & 33.6          \\
\midrule

\multicolumn{4}{c}{\textbf{OpenEQA Subset}}                                                                 \\ \midrule
\multicolumn{1}{c|}{\textbf{Method}}                  & \textbf{ScanNet} & \textbf{HM3D}  & \textbf{ALL}   \\
\midrule
\multicolumn{1}{c|}{Video-LLaVA}                       & 32.9            & 27.8          & 30.6          \\
\multicolumn{1}{c|}{LLaMA-VID}                       & 31.2            & 28.0          & 29.4          \\
\multicolumn{1}{c|}{VideoAgent}                       & 37.6            & 34.6          & 36.3          \\
\multicolumn{1}{c|}{\textbf{E-VideoAgent(InternVL2-8B)}} & 39.7            & 43.0          & 41.2          \\
\multicolumn{1}{c|}{\textbf{E-VideoAgent(GPT-4o)}}    & \textbf{46.0}   & \textbf{48.2} & \textbf{47.0} \\ \bottomrule
\end{tabular}}
\label{openeqa_res}
\vskip -0.1in
\end{table}

\begin{table}[]
\caption{Results of open-ended question answering over embodied interactions on the test set of EnvQA~\cite{envqa}.}
\resizebox{\columnwidth}{!}{
\begin{tabular}{@{}cccc@{}}
\toprule
\multicolumn{4}{c}{\textbf{EnvQA}}                                                                      \\ \midrule
\multicolumn{1}{c|}{\textbf{Method}}              & \textbf{Events} & \textbf{Orders} & \textbf{States} \\ \midrule
\multicolumn{1}{c|}{Video-LLaVA}                  & 10.19           & 39.00           & 18.50           \\
\multicolumn{1}{c|}{LLaMA-VID}                    & 9.98            & 54.00           & 5.50            \\
\multicolumn{1}{c|}{VideoAgent}                    & 5.54            & 65.5           & 12.5           \\
\multicolumn{1}{c|}{\textbf{Embodied VideoAgent}} & \textbf{25.91}           & \textbf{68.00}           & \textbf{35.50}           \\ \bottomrule
\end{tabular}
}
\label{envqa_res}
\vskip -0.2in
\end{table}

\begin{figure*}[t!]
    \centering
    \includegraphics[width=\textwidth]{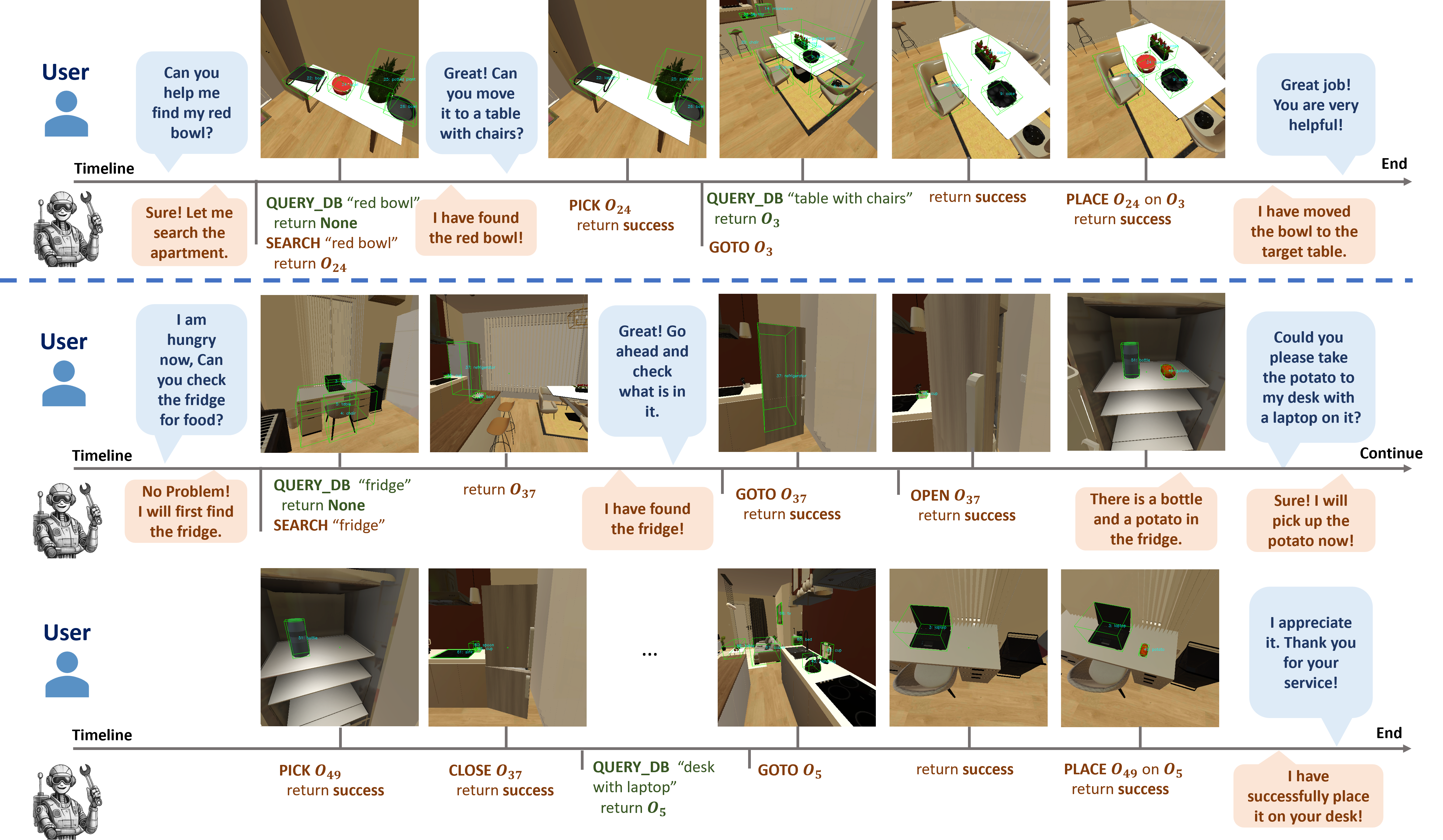}
    \caption{An episode of generated embodied user-assistant interaction. The episode is produced by the framework mentioned in \autoref{sec:two_agent}, where an LLM plays the user and \method is the assistant. The episode comprises various embodied problem-solving that requires precise memory of the scene objects and tool usage. More example episodes can be found in \supp.}
    \label{fig:two_agent_examples}
    \vskip -0.2in
\end{figure*}

\begin{figure*}[t]
    \centering
    \includegraphics[width=\textwidth]{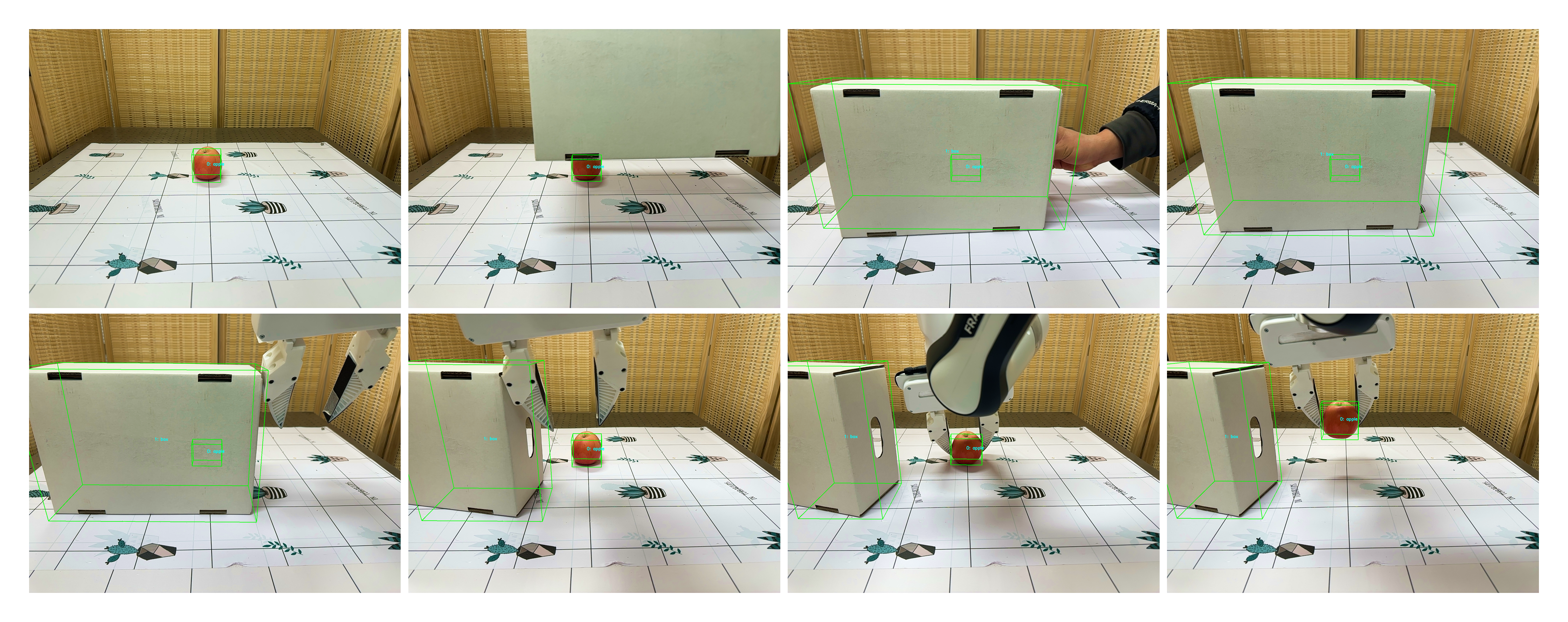}
    \caption{Our persistent object memory enables effective real-world robotic manipulation. Using \method for perception, the robot is tasked to pick up an apple, which soon becomes occluded by a box. Leveraging its memory, the robot retrieves the apple’s position, moves the box aside, and successfully completes the task.}
    \label{fig:manipulation_demo}
    \vskip -0.2in
\end{figure*}

\autoref{openeqa_res} shows the results. It can be inferred that OpenEQA Subset is harder than the full OpenEQA validation set from the performance drops of Video-LLaVA and LLaMA-VID. On the hard subset, the two variants of \method both achieve good performances. Specifically, \method(GPT-4o) obtained $46.0\%$ on ScanNet and $48.2\%$ on HM3D, surpassing Video-LLaVA by $13.1\%$ and $20.4\%$ on the ScanNet and HM3D respectively. 

\noindent
\textbf{Temporal localization + \toolD\xspace\texttt{vqa} tool solves embodied questions better than scene graph.~~} The performance gaps between GPT4-w/LLaVA-1.5 and GPT-4 w/CG on the full OpenEQA indicate that LLM can better utilize frame captions for question answering other than scene graphs, which is validated by the better performance of \method ($+16.4\%$ over Video-LLaVA on subset) over GPT-4 w/CG ($+4.4\%$ over Video-LLaVA on full set). \method does not explicitly construct a complex scene graph during memory construction. Instead, for relational questions about two objects during the inference, \method will use temporal localization to retrieve the frame that contains both the objects and use VLM to answer the relational questions about them. We found that the strong performance of \method mainly attributed to the precise frame localization using consistent object memory and history buffer.

\noindent
\textbf{Agentic systems outperforms End-to-End VLMs.~~} On the OpenEQA subset, agentic methods \method(InternVL2-8B), \method(GPT-4o) and VideoAgent all achieve better results than end-to-end models Video-LLaVA and LLaMA-VID due to their multi-step information retrieval and reasoning abilities. Besides, the performance gains of \method over VideoAgent suggest that a consistent object memory with comprehensive features (object feature, object context feature, and frame feature)
will leads to better temporal and spatial localization, which finally leads to accurate question answering.

\subsubsection{Results on EnvQA}\label{sec:exp_envqa}
The results on EnvQA are shown in \autoref{envqa_res}. \method achieves a significant performance gain compared to its three counterparts.

\noindent
\textbf{VLM-based memory update plays a key role in event understanding.} The crucial component for accurately answering questions about Events and Orders is the action buffer presented in \method, which associates each action with its target object. The difficulty in understanding events in EnvQA is the absence of ``hand'' in the simulated environment, which makes the action annotator less effective. With the help of VLM-based memory update, \method, by contrast, can better identify the critical target objects in the dynamic scenes, resulting in better performance. 

\noindent
\textbf{Object relation detection helps to solve States question.} The States questions involve recalling the final position of an object. By automatic object relation detection using 3D bbox, the final receptacle that holds the object can be retrieved from the \textbf{RO} field, therefore enhancing the ability of \method to answer States questions.

\subsection{More Applications in Embodied AI Tasks}\label{sec:exp_qualitative}
We further explore the potential of \method across various embodied AI tasks. In \autoref{fig:two_agent_examples}, we illustrate an episode of user-assistant interaction generated by the two-agent framework described in \autoref{sec:two_agent}, all within the AI-Habitat simulator~\cite{savva2019habitat}. As shown, to enable such interactions, the assistant (powered by \method) requires an accurate and comprehensive understanding of scene objects it has previously encountered, such as “my desk with a laptop on it.” \method effectively fulfills diverse requests from the LLM user by seamlessly integrating memory query tools with embodied action primitives. Additional examples of these embodied interactions are provided in \supp. In \autoref{fig:manipulation_demo}, we showcase \method’s application in robotic perception, where a Franka robot uses it to build persistent memory in a dynamic manipulation scene. In this task, the robot is instructed to pick up an apple. However, the apple later becomes hidden behind a box, illustrating the dynamic nature of the scene. Leveraging persistent object memory, the robot successfully recalls the apple's location despite the obstruction and completes the task by first moving the box aside, demonstrating the effectiveness of scene memory.

\section{Related Works}

\noindent
\textbf{Video and Dynamic Scene Understanding.~~} Most existing 3D scene understanding methods struggle with dynamic scenes due to limitations in input modalities, as both 2D images and 3D point clouds inherently capture static information~\cite{gu2024conceptgraphs,jatavallabhula2023conceptfusion}. To remedy this, dynamic scene understanding is introduced to facilitate this~\cite{worldqa,mangalam2024egoschema,xiao2021next,zhang2023video,maaz2023video}.  Unlike canonical scene understanding, which primarily focuses on identifying static objects and entities, dynamic scene understanding centers on how actions and activities affect these elements within a scene. Research in this area typically uses video as the primary modality due to its natural ability to capture dynamic changes and its relative ease of acquisition~\cite{ma2023sqa3d,openeqa}. Moreover, dynamic scene understanding often involves long-form video~\cite{wang2024lstp,videoagent}, adding layers of complexity. Recent approaches explore egocentric video~\cite{ego4d,mangalam2024egoschema,Damen2022RESCALING} and large-scale multimodal training~\cite{pramanick2023egovlpv2,lin2022egocentric,videollava,llamavid,wang2024lstp}. Despite these advancements, existing models struggle with performance issues due to the inherent complexity of the task~\cite{worldqa,openeqa,ego4d} or face high computational demands~\cite{team2023gemini,team2024gemini}, limiting their applicability for embodied agents and robotic systems.

\noindent
\textbf{Multimodal Agents for Perception.~~}Recent advancements in large language models (LLMs) have showcased impressive reasoning and problem-solving abilities across diverse domains~\cite{openai2023gpt4,team2023gemini,team2023gemini,dubey2024llama}, leading to their application in perception tasks~\cite{gao2023clova,videoagent,yang2024doraemongpt,suris2023vipergpt,gupta2023visual}. This approach leverages LLMs to decompose complex perception tasks (such as visual question answering) into smaller, manageable subtasks. These subtasks are then completed by multiple specialized tool models (often end-to-end models), and the outputs are aggregated by the LLM into a cohesive response~\cite{gao2023clova,videoagent,gupta2023visual}. Known as \textit{multimodal agents}, these systems have shown promising results in 2D image comprehension~\cite{gao2023clova}, video analysis~\cite{videoagent,yang2024doraemongpt}, and 3D scene understanding~\cite{huang2024chat}. Compared to traditional end-to-end methods, agent-based perception offers reduced training and inference costs, improved explainability through explicit chain-of-thought reasoning, and, in some cases, faster inference rates (\eg, for long-form video understanding). However, current methods still face challenges with more complex tasks, such as dynamic scene understanding, likely due to limitations in the flexibility and precision of memory design.

\section{Conclusions}

We have presented \method, a memory-augmented multimodal tool-use agent that tackles the challenging dynamic scene understanding tasks with a novel persistent object memory and an automatic memory update method based on VLMs. Compared to end-to-end multimodal LLMs and tool-use agent counterparts, the memory architecture of \method enables precise, comprehensive scene understanding by integrating egocentric observations with embodied sensory inputs (e.g., depth maps, camera poses). This design is resilient to ongoing changes in the scene caused by various actions and activities performed by embodied agents, making it particularly well-suited for embodied AI tasks. The effectiveness of \method has been validated through the promising results on various embodied scene understanding tasks including Ego4D-VQ3D, OpenEQA, and EnvQA. Future directions may involve deploying robots in more challenging environments, such as production sites and outdoor settings.

{
    \small
    \bibliographystyle{ieeenat_fullname}
    \bibliography{main}
}


\twocolumn[{%
\renewcommand\twocolumn[1][]{#1}%
\newpage
}]
\appendix
\section{Fields of the Object Entry}
An object in Persistent Object Memory has the following fields:
\begin{itemize}
    \item \textbf{ID}: The unique object ID in the memory, together with the detected category. Our 3D object re-identification algorithm can be found in \autoref{sec:re-id_detail}.
    
    \item \textbf{STATE}: The object state can be "open", "close", "in hand" or "normal". It is updated by VLM, which will be discussed in \autoref{app_vlm_update}.
    \item \textbf{Related Objects(RO)}: A list of objects that have "on", "uphold", "in" and "contain" relations with the entry object. The detections of These relations are based on 3D bounding boxes. For example, Given the 3D bounding boxes $B_1$ and $B_2$ of object $O_1$ and $O_2$ correspondingly, if $B_1$ has a higher altitude than $B_2$, $B_1$ has contact with $B_2$ and $B_1$ is inside the horizontal surface of $B_2$, then $O_1$ is "on" $O_2$ and $O_2$ "upholds" $O_1$.
    \item \textbf{3D Bbox}: It is obtained by 2D-3D lifting and dynamically updated by the moving average algorithm. Please refer to \autoref{sec:2d-3d-lifting} and \autoref{sec:re-id_detail} for more details.
    \item \textbf{OBJ Feat}: It is the CLIP feature of the object's cropped image. It is updated by the moving average algorithm. Details are provided in \autoref{sec:re-id_detail}.
    \item \textbf{CTX Feat}: It is the CLIP feature of the frame where the object is visible. It is updated by the moving average algorithm. Please refer to \autoref{sec:re-id_detail} for details.
\end{itemize}

\section{2D-3D Lifting}
\label{sec:2d-3d-lifting}
In this paper, 2D-3D Lifting refers to getting the 3D bounding boxes of the objects using 2D object detection bounding boxes, camera poses, and depth images. Different from methods that use point clouds or voxels to represent 3D object geometry, we found that representing object geometry as 3D bounding boxes is enough for 
embodied perception. More importantly, compared to point clouds or voxels, 3D bounding boxes are more memory-efficient and can be maintained and updated easily, which makes them a natural choice for 3D object perception in dynamic scenes. 

To get the 3D bounding boxes of the objects, we first use YoloWorld\cite{yoloworld} detector to predict the 2D bounding boxes of the objects. SAM-2\cite{sam2} is then adopted to get the corresponding object masks for the detected objects given the frame and the bounding boxes. For each object, we use its 2D object mask to get its depth pixels and transform them into object surface points in the world coordinate system using the camera intrinsic and extrinsic. We then filter out the bad points (usually, the foreground and the background pixels caused by imperfect segmentation mask prediction) by simply sorting the object surface points by their distances to the camera, and removing the first $10\%$ and the last $10\%$ of the points to finally get the refined object surface points. The object bounding boxes are then computed based on the minimum and maximum values of the points' coordinates.

\begin{algorithm*}[t] 
    \caption{\label{alg:static_reid}Static Object Re-Identification.}  
    \KwIn {detected object $O_k$, static object list $\mathcal{S}=[S_1, S_2, ..., S_m]$}
    \KwOut {re-IDed object if $O_k$ matches one of the static objects else $O_{k}$}
    \For {$S_i$ in $[S_1, S_2, ...S_n]$}{
        \If {$\text{Spatial\_IoU}(O_k, S_i) > 0.2$ \textbf{or} ($\text{Spatial\_MaxIoS}(O_k, S_i) > 0.2$ \textbf{and} ${O_k}.\text{category}$ == ${S_i}.\text{category}$)}
        {
         return \textbf{True}, $S_i$
         }
    }
    return \textbf{False}, $O_k$
\end{algorithm*}
\begin{algorithm*}[t] 
    \caption{\label{alg:dynamic_reid}Dynamic Object Re-Identification.}  
    \KwIn {detected object $O_k$, dynamic object list $\mathcal{D}=[D_1, D_2, ..., D_n]$}
    \KwOut {re-IDed object if $O_k$ matches one of the dynamic objects else $O_k$}
    \For {$D_i$ in $[D_1, D_2, ...D_n]$}{
        \If {$\text{Spatial\_Vol\_Sim}(O_k, D_i) > 0.7$ \textbf{and} $\text{Visual}(O_k, D_i) > 0.45$}
        {
         return \textbf{True}, $D_i$
         }
    }
    return \textbf{False}, $O_k$
\end{algorithm*}

\begin{algorithm*}[t] 
    \caption{\label{alg:reid}Object Memory Update.}  
    \KwIn {current observations $\text{Obs}^t=\{\text{RGB}^t, \text{Depth}^t, \text{Pose}^t\}$, previous object memory  $\mathcal{M}_O^{t-1}$}
    \KwOut {current object memory $\mathcal{M}_O^t$}
    $\text{2DBoxes}, \text{categories} = \text{2D\_Detector}(\text{RGB}^t)$ \\
    $\mathcal{S}, \mathcal{D} = \text{ObjectSplit}(\mathcal{M}_O^{t-1}, \text{Obs}^t)$ ~~~//See \cref{sec:object_split} \\
    \For {\textnormal{$i$ in range(len(2DBoxes))}}{
        
        $\text{category} = \text{categories}[i]$ \\
        $\text{2DBox} = \text{2DBoxes}[i]$ \\
        $\text{3DBox} = \text{2D\_3D\_Lifting}(\text{2DBox}, \text{Obs}^t)$~~~//See \cref{sec:2d-3d-lifting} \\
        $\text{Feat}_{\text{CLIP}} = \text{CLIP\_Model}(\text{RGB}^t[\text{2DBox}])$\\
        $\text{Feat}_{\text{DINOv2}} = \text{DINOv2\_Model}(\text{RGB}^t[\text{2DBox}])$\\
        $O_{\text{tmp}} = \text{Object3D}(\text{category}, \text{3DBox}, \text{Feat}_{\text{CLIP}}, \text{Feat}_{\text{DINOv2}})$ \\
        $\text{sgn}, O_{\text{ID}} = \text{Static\_Object\_ReID}(O_{\text{tmp}}, \mathcal{S})$ ~~~/first try to re-identify $O_{\text{tmp}}$ from static objects (\autoref{alg:static_reid}) \\
        \If {\textnormal{\text{sgn} == \textbf{True}}}
        {
            $O_{\text{ID}} = \text{Static\_Object\_Merge}(O_{\text{tmp}}, O_{\text{ID}})$ \\
        }
        \Else 
        { 
            $\text{sgn}, O_{\text{ID}} = \text{Dynamic\_Object\_ReID}(O_{\text{tmp}}, \mathcal{D})$ ~~~//try to re-identify $O_{\text{tmp}}$ from dynamic objects (\autoref{alg:dynamic_reid})\\
            \If {\textnormal{\text{sgn} == \textbf{True}}}
            {
                $O_{\text{ID}} = \text{Dynamic\_Object\_Merge}(O_{\text{tmp}}, O_{\text{ID}})$ \\
                move $O_{\text{ID}}$ from $\mathcal{D}$ to $\mathcal{S}$    
            }
            \Else{
                add $O_{\text{tmp}}$ to $\mathcal{S}$ ~~~//$O_{\text{tmp}}$ is a brand new object\\
                }
        }
    }
    $\mathcal{M}_O^t = \mathcal{S} \cup \mathcal{D}$\\
    $\mathcal{M}_O^t = \text{Related\_Object\_Update}(\mathcal{M}_O^t)$\\
    $\mathcal{M}_O^t = \text{VLM\_Update}(\mathcal{M}_O^t, \text{RGB}^t)$\\
    return $\mathcal{M}_O^t$
\end{algorithm*}

\section{Object Re-Identification in Dynamic Scenes}
\label{sec:re-id_detail}
Accurate object re-identification (re-ID) in dynamic scenes can better facilitate embodied perception, task planning, and reasoning. \method utilizes both object visual features and object 3D bounding boxes for object re-identification. The visual similarity score and the spatial similarity score of an object pair are detailed as follows.

\subsection{Visual Similarity Score}
For a detected object on the 2D frame, we crop the object image from the frame using its 2D bounding box and extract the CLIP\cite{radford2021learning} and DINOv2\cite{oquab2023dinov2} 
features of this image crop as the object's visual features. To calculate the visual similarity of two objects, we use the following visual similarity score\cite{videoagent}:
\begin{equation}
\resizebox{0.9\linewidth}{!}{
    $ \text{Visual}(O_i, O_j) = 0.15*\text{CLIP}(O_i, O_j)+0.85*\text{DINOv2}(O_i, O_j)$
}
\end{equation}
where $\text{Visual}(O_i, O_j)$ denotes visual similarity of object $O_i$ and $O_j$, 
$CLIP(\cdot, \cdot)$ and $DINOv2(\cdot, \cdot)$ are the CLIP and DINOv2 similarities proposed in \cite{videoagent}.

Besides the CLIP feature of the cropped object image (denoted as \textbf{OBJ Feat} in \autoref{fig:method}), the CLIP feature of the frame containing the object is also stored as the context feature of the object (denoted as \textbf{CTX Feat} in \autoref{fig:method}). The context feature will not be used for object re-ID, but it later enables retrieving objects by an open-vocabulary environment description ("blue wall", "kitchen", etc) during inference.

\subsection{Spatial Similarity Scores}
Given two objects $O_1$ and $O_2$ and their 3D bounding boxes: $[[x_1^{min}, y_1^{min}, z_1^{min}], [x_1^{max}, y_1^{max}, z_1^{max}]]$ and $[[x_2^{min}, y_2^{min}, z_2^{min}], [x_2^{max}, y_2^{max}, z_2^{max}]]$, their volumes and the volume of their intersection can be easily computed as:
\begin{align*}
V_1 &= (x_1^{max}-x_1^{min})(y_1^{max}-y_1^{min})(z_1^{max}-z_1^{min}), \\
V_2 &= (x_2^{max}-x_2^{min})(y_2^{max}-y_2^{min})(z_2^{max}-z_2^{min}), \\
x_{inter} &= min(x_1^{max}, x_2^{max})-max(x_1^{min}, x_2^{min}), \\
y_{inter} &= min(y_1^{max}, y_2^{max})-max(y_1^{min}, y_2^{min}), \\
z_{inter} &= min(z_1^{max}, z_2^{max})-max(z_1^{min}, z_2^{min}), \\
V_{inter} &= max(0, x_{inter})*max(0, y_{inter})*max(0, z_{inter}),\\
V_{union} &= V1+V2-V_{inter}
\end{align*}
where $V_1$ and $V_2$ are the volumes of $O_1$ and $O_2$, $V_{inter}$ is the volume of their intersection and $V_{union}$ is the volume of their union.
we use three scores to evaluate the similarity of the two bounding boxes:

Intersection over Union \textbf{(IoU)}:
\begin{equation}
\text{Spatial\_IoU}(O_i, O_j) = \frac{V_{inter}}{V_{union}}.
\end{equation}

Maximum Ratio of Intersection over Subsets \textbf{(MaxIoS)}:
\begin{equation}
\text{Spatial\_MaxIoS}(O_i, O_j) = max(\frac{V_{inter}}{V_1}, \frac{V_{inter}}{V_2}).
\end{equation}

Bounding Box Volume Similarity \textbf{(Vol\_Sim)}
\begin{equation}
\text{Spatial\_Vol\_Sim}(O_i, O_j) = \frac{min(V_1, V_2)}{max(V_1, V_2)}.
\end{equation}
These three scores evaluate object spatial proximity from three different perspectives: 
\begin{itemize}
\item \textbf{$\text{Spatial\_IoU}$}: When two bounding boxes have similar volumes and have large intersection volume, $\text{Spatial\_IoU}$ will approach its maximum value $1$. It is a strong indicator (when $\text{Spatial\_IoU}>0.2$) of two bounding boxes referring to the same object.

\item \textbf{$\text{Spatial\_MaxIoS}$}: When two bounding boxes demonstrate a strong containment relationship,  $\text{Spatial\_MaxIoS}$ will get closer to its maximum value $1$. For example, given that $O_1$ and $O_2$ are both detected as 'table', $O_2$ is $\frac{1}{10}$ the volume of $O_1$ and its bounding box is inside $O_1$, $\text{Spatial\_MaxIoS}$ will reach $1$, while their $\text{Spatial\_IoU}$ is only $0.1$. It is used together with object 
categories to re-identify partially observed objects due to occlusion. In the above example, $O_2$ is possibly a partial observation of $O_1$ given that they have overlapping bounding boxes and the same object category.

\item \textbf{$\text{Spatial\_Vol\_Sim}$}: when two bounding boxes have similar volume, $\text{Spatial\_Vol\_Sim}$ will have larger value. It is used along with visual similarity scores to match dynamic objects.
\end{itemize}

\subsection{Recognizing Dynamic Objects}
\label{sec:object_split}
With the knowledge of both object visual features and 3D bounding boxes, we can perform object re-identification based on both visual similarity and spatial similarity. For static objects, spatial similarity serves as a valuable metric for object re-ID. However, for dynamic objects, object re-ID should focus more on the visual similarity of the object pairs, since the object positions are dynamically changing. Therefore, before re-identifying the newly detected objects, we should first classify the existing objects in the object memory into static objects and dynamic objects.

The key idea of recognizing dynamic objects in the object memory is straightforward: if an object is not where it should be, then it must be moved by someone (becomes dynamic). We first retrieve the objects from the object memory whose 3D bounding boxes can be directly viewed on the current frame (achieved by world-to-camera transformation) with no occlusion (achieved by validating the depth values of the corresponding pixels). For each retrieved object, We then compare the visual features of "where it should be" on the current frame with its visual features in the object memory. If the visual similarity score is below a threshold (0.45 in our settings), then the object is not "where it should be" and should be marked as "dynamic". By this method, before performing object-reID on current detections, we split the objects in the object memory into two sets: static objects $\mathcal{S}$ and dynamic objects $\mathcal{D}$.

\subsection{Object Re-ID for Static and Dynamic Objects}
\label{sec:object_reid}
\autoref{alg:static_reid} and \autoref{alg:dynamic_reid} are the object re-ID methods for static objects and dynamic objects correspondingly. Each algorithm receives a newly detected object $O_k$ with visual features and its 3D bounding box, and a list of candidate objects (static object list or dynamic object list). They both return whether the object $O_k$ can be successfully identified and the object ID of the matched object in the candidate list. If $O_k$ is re-identified, it is merged into the matched object by performing a moving average on the fields of the 3D bounding box and visual features. Specifically, to merge the two objects matched by static object re-ID, the window size of the moving average is set to $10$, leading to a mild change in object visual features and spatial occupation; for dynamic object merging, we set the window size to $2$, allowing rapid change of visual features and bounding boxes due to object movement.

\autoref{alg:reid} presents an overview of object memory update, including 3D object detection and re-ID. The main idea is to first divide the objects in $\mathcal{M}_O^{t-1}$ into static ones $\mathcal{S}$ and dynamic ones $\mathcal{D}$, and try to match the newly detected objects to these two kinds of objects through \autoref{alg:static_reid} and \autoref{alg:dynamic_reid} respectively. If successfully matched, the newly detected objects will be merged with the matched objects in the object memory using the moving average as mentioned, otherwise, it will be viewed as a brand new object and added to the object memory. Finally, VLM-based Memory update will be performed on $\mathcal{M}_O^t$, which will be discussed in \autoref{app_vlm_update}.

\section{VLM-based Memory Update}
\label{app_vlm_update}
When \method serves as an observer of an egocentric video, \method needs to predict the actions of the camera wearer in the video and associate the object IDs in the object memory with the subjects of the actions. We use LaViLa\cite{lavila} to annotate the action of the camera wearer every two seconds. For each action annotation, we first prompt an LLM (GPT-4o) to extract the objects in the annotation (e.g. "bottle" and fridge" given the annotation "\#C C picks the bottle from the fridge") and select candidate objects detected at that time according to their categories for matching. We then perform VLM-based object association illustrated in \autoref{fig:vlm_update}, and save the actions to Action Buffer. Finally, we query the state change of the matched objects and update the "STATE" field of the object entries. In this paper, objects have one of the following states: "open", "close", "in hand" and "normal".

When \method is equipped with embodied actions, the procedure of VLM-based object association is omitted since \method serves as an active planner with the knowledge of the object IDs of its target objects or receptacles. In this case, VLM serves as an action validator that judges whether an action is successfully performed and updates the "STATE" field of the target objects.

\section{Results under Noisy Camera Poses}
We conduct the ablation study of the influence of the noisy camera poses. On OpenEQA benchmark, We provide \method(InternVL-2) with 1) the accurate camera poses provided in habitat simulator, denoted as \textbf{E-VideoAgent(GT poses)}; 2) the estimated camera poses and depths via DUSt3R\cite{wang2024dust3r}, denoted as \textbf{E-VideoAgent(noisy poses)}. Results in \autoref{openeqa_noisy_res} show that \method can also handle perception tasks well based on the noisy poses, suffering little performance drops when using the estimated camera poses and depths. This suggests further applications of \method on RGB videos only, with the camera poses and depths being estimated by cutting-edge scene reconstruction methods.  

\begin{table}[]
\caption{Results of \method under noisy poses.}
\resizebox{\columnwidth}{!}{
\begin{tabular}{@{}cccc@{}}
\toprule
\multicolumn{4}{c}{\textbf{OpenEQA Subset}}                                                                 \\ \midrule
\multicolumn{1}{c|}{\textbf{Method}}                  & \textbf{ScanNet} & \textbf{HM3D}  & \textbf{ALL}   \\
\midrule
\multicolumn{1}{c|}{Video-LLaVA}                       & 32.9            & 27.8          & 30.6          \\
\multicolumn{1}{c|}{LLaMA-VID}                       & 31.2            & 28.0          & 29.4          \\
\multicolumn{1}{c|}{VideoAgent}                       & 37.6            & 34.6          & 36.3          \\
\multicolumn{1}{c|}{\textbf{E-VideoAgent(GT poses)}} & \textbf{39.7}            & \textbf{43.0}          & \textbf{41.2}          \\
\multicolumn{1}{c|}{\textbf{E-VideoAgent(noisy poses)}}    & 38.2   & 42.2 & 40.0 \\ \bottomrule
\end{tabular}}
\label{openeqa_noisy_res}
\vskip -0.1in
\end{table}

\section{Embodied Perception}
For embodied perception, we equip \method with the following tools:
\begin{itemize}
    \item \toolC\texttt{query\_db}: Given a query, this tool will return the candidate object entries from Persistent Object Memory. It is a combination of code-based retrieval (writing a piece of MySQL code to query the database) and similarity-based retrieval. For similarity-based retrieval, \toolC\texttt{query\_db} supports 
    \texttt{retrieve\_objects\_by\_appearance} (based on text-image similarities between the query text and the \textbf{OBJ Feats}) and \texttt{retrieve\_objects\_by\_environment} (based on text-image similarities between the query text and the \textbf{CTX Feats}).
   
    \item \toolA\texttt{temporal\_loc}: Return the top-5 frame IDs that satisfy the description (e.g. when I walk in the front door). It is achieved by the text-image similarity between the input description and the frame features stored in the temporal memory $\mathcal{M}_T$.
    
    \item \toolB\texttt{spatial\_loc}: Return the top-3 3D positions that satisfy the description (e.g. bedroom). It is achieved by calculating the center positions of the top-3 object spatial clusters where objects have strong \textbf{CTX feat} similarities to the input text description. This is only used for embodied navigation.
    
    \item \toolD\texttt{vqa}: Given an image (can be a video frame, a cropped object image, or a frame plotted with a 3D bounding box referring to a specific object), this tool will describe the image and then answer the question.
\end{itemize}

We use the following prompt for perception tasks, with \{\textbf{tools}\} in the prompt being the above tools. We choose GPT-4o as the LLM agent and InternVL2 as the VLM for visual question answering.

\begin{tcolorbox}[breakable]
You are tasked with answering a question about a scene. There is a SQL database that contains the following tables:

TABLE Objects(
    
    object\_id INT,
    
    category VARCHAR(255),
    
    volume FLOAT,
    
    PRIMARY KEY (object\_id)

)

TABLE Objects\_Frames(

    object\_id INT,

    frame\_id INT

)

The 'Objects' table catalogues the objects detected in the scene with their categories and volumes. The categories are from \{\textbf{categories\_list}\}. The 'Objects\_Frames' table records the appearing objects in the frames.
~\\

You have access to the following tools:
~\\

\{\textbf{tools}\}
~\\

ATTENTION: 

1. Use double quotes for all string arguments when using tools, avoiding any single quotes. If the tool has two arguments, format the output within brackets, like this: ("What is the color of the object in the image?", 16).

2. Prioritize using retrieve\_objects\_by\_appearance, retrieve\_objects\_by\_environment, and frame\_localization for object and frame localization, rather than query\_database.

3. Once the relevant frames are located, use frame\_VQA to ask questions about these frames.

4. For "where" questions, interpret them as spatial location queries, not as requests to identify specific frames. Use frame\_VQA to provide these answers.

5. If the information is insufficient for a precise response, generate a response based on the available data.
~\\

Use the following format:
~\\

Question: the input question you must answer

Thought: you should always think about what to do

Action: the action to take, should be one of [\{\textbf{tool\_names}\}]

Action Input: the input to the action

Observation: the result of the action... (this Thought/Action/Action Input/Observation can repeat N times)

Thought: I now know the final answer

Final Answer: the answer to the original input question
~\\

Begin!
~\\

Question: \{\textbf{input}\}

Thought: \{\textbf{agent\_scratchpad}\}

\end{tcolorbox}

\twocolumn[{%
\renewcommand\twocolumn[1][]{#1}%
\subsection{Example 1}
\label{sec_success_1}
\begin{itemize}
    \item \textbf{Question}: "What color are the blinds?"
    \item \textbf{Ground-Truth Answer}: "White"
    \item \textbf{Predicted Answer}: "The blinds are white."
\end{itemize}
In the following Chain Of Thought (CoT), \method first successfully locates the frames containing the target object "blinds" by  "frame\_localization" as shown in \autoref{success_1}. Then, \method calls the VQA module (InternVL2) to first describe the located frame and then answer the question based on this frame, and finally gets the correct answer: "The blinds are white".

\begin{tcolorbox}[breakable]
Entering new AgentExecutor chain...

\textbf{Thought}: To determine the color of the blinds, I need to locate frames that might contain blinds and then analyze those frames.

\textbf{Action}: frame\_localization

\textbf{Action Input}: "blinds"

\textbf{Result}: The most relevant frame indices are [190, 200, 210, 180, 240] 

\textbf{Thought}: I have identified the frames that likely contain blinds. Now, I will use frame\_VQA to determine the color of the blinds in these frames.

\textbf{Action}: frame\_VQA

\textbf{Action Input}: ("What is the color of the blinds?", 190)

\textbf{Result}: Description: The image shows an indoor scene featuring a window with partially open horizontal blinds. Below the window is a heater or radiator, and part of a stove is visible on the left side. Outside the window, some greenery can be seen.

Answer to the question: The blinds are white.

\textbf{Thought}: I now know the final answer.

\textbf{Final Answer}: The blinds are white.

Finished chain.
\end{tcolorbox}

\begin{center}
    \includegraphics[width=0.6\linewidth]{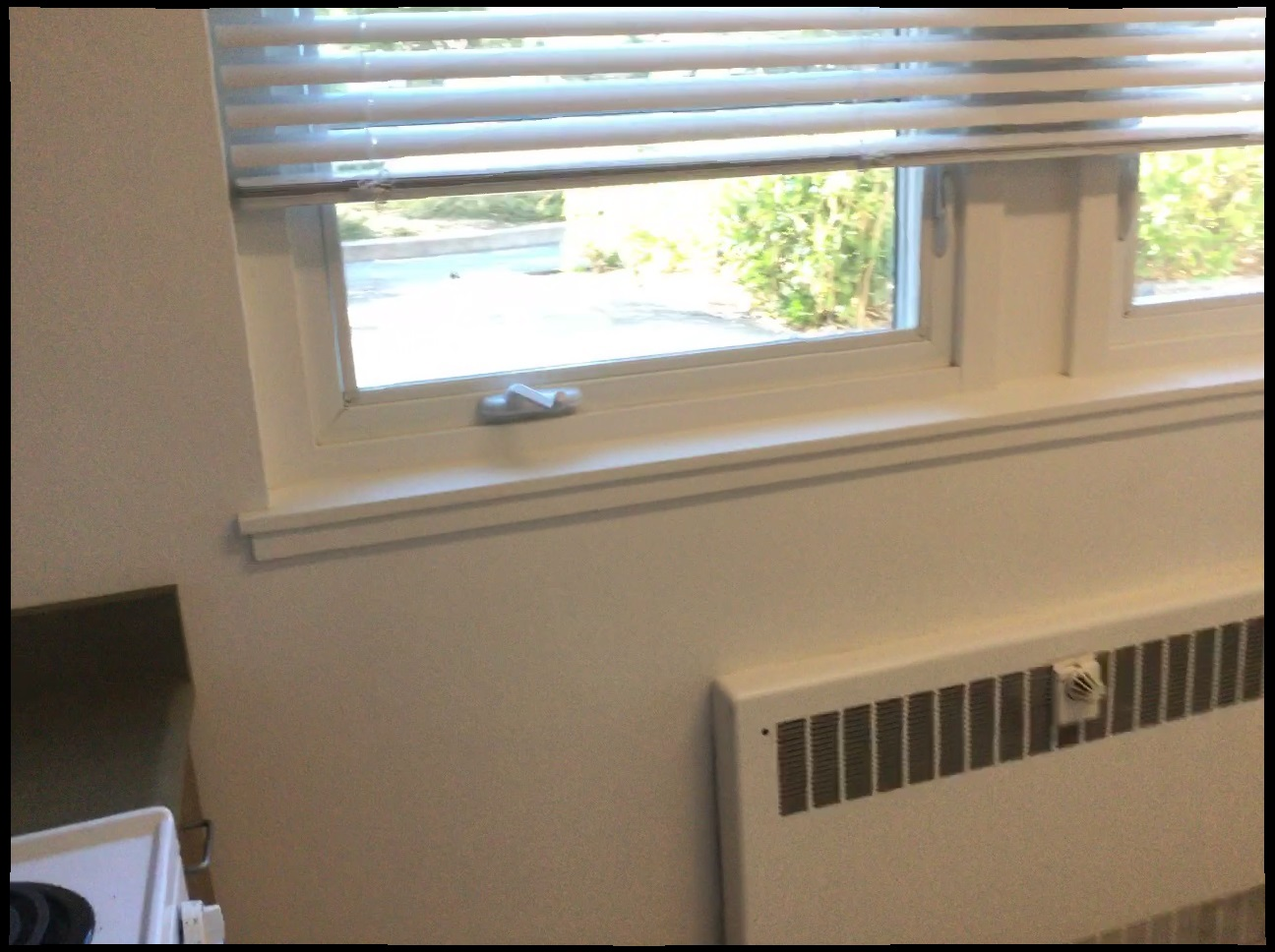}
    \captionof{figure}{Frame No.190 that contains blinds are successfully located by \method in \autoref{sec_success_1}.}
    \label{success_1}
\end{center}
}]

\twocolumn[{%
\renewcommand\twocolumn[1][]{#1}%
\subsection{Example 2}
\label{sec_success_2}
\begin{itemize}
    \item \textbf{Question}: "How many cups does the Mr coffee make?"
    \item \textbf{Ground-Truth Answer}: "12"
    \item \textbf{Predicted Answer}: "The Mr. Coffee machine makes 12 cups"
\end{itemize}
To answer this question, \method uses temporal localization to locate the frame containing the coffee machine as illustrated in \autoref{success_2}. Then, it queries InternVL2 for a detailed description of this coffee machine. InternVL2 successfully identifies the words on the box and finally answers that Mr. Coffee is capable of making 12 cups. The CoT for the inference is provided as follows.

\begin{tcolorbox}[breakable]
Entering new AgentExecutor chain...

\textbf{Thought}: To answer the question about how many cups the Mr. Coffee makes, I need to locate frames that might show the Mr. Coffee machine and then analyze those frames to determine its capacity.

\textbf{Action}: frame\_localization

\textbf{Action Input}: "Mr. Coffee machine"

\textbf{Result}: The most relevant frame indices are [270, 280, 260, 250, 290]

\textbf{Thought}: I have identified the frames that likely contain the Mr. Coffee machine. I will now use the frame\_VQA tool to analyze these frames and determine the capacity of the Mr. Coffee machine.

\textbf{Action}: frame\_VQA

\textbf{Action Input}: ("How many cups does the Mr. Coffee make?", 270)

\textbf{Result}: Description: The image shows a kitchen section with multiple shelves. 

- On the topmost shelf, there is a white box labeled “Hamilton Beach” with visible text indicating that it is related to a kitchen appliance, and a photo of the appliance is shown on the box. This section includes other miscellaneous items.

- The middle shelf contains a big brown box labeled “Mr. Coffee.” The box features a picture of a coffee maker.

- In the bottom shelf, there are two colorful sponges – one red and one green – along with various other items that can't be clearly identified.

This portion of the room appears to be organized and neatly arranged, with items stored in an accessible manner.

 Answer to the question: Based on the label visible in the image, the Mr. Coffee maker is capable of making 12 cups.
 
\textbf{Thought}: I have determined the capacity of the Mr. Coffee machine from the frame description.

\textbf{Final Answer}: The Mr. Coffee machine makes 12 cups.

Finished chain.
\end{tcolorbox}
\begin{center}
    \includegraphics[width=0.6\linewidth]{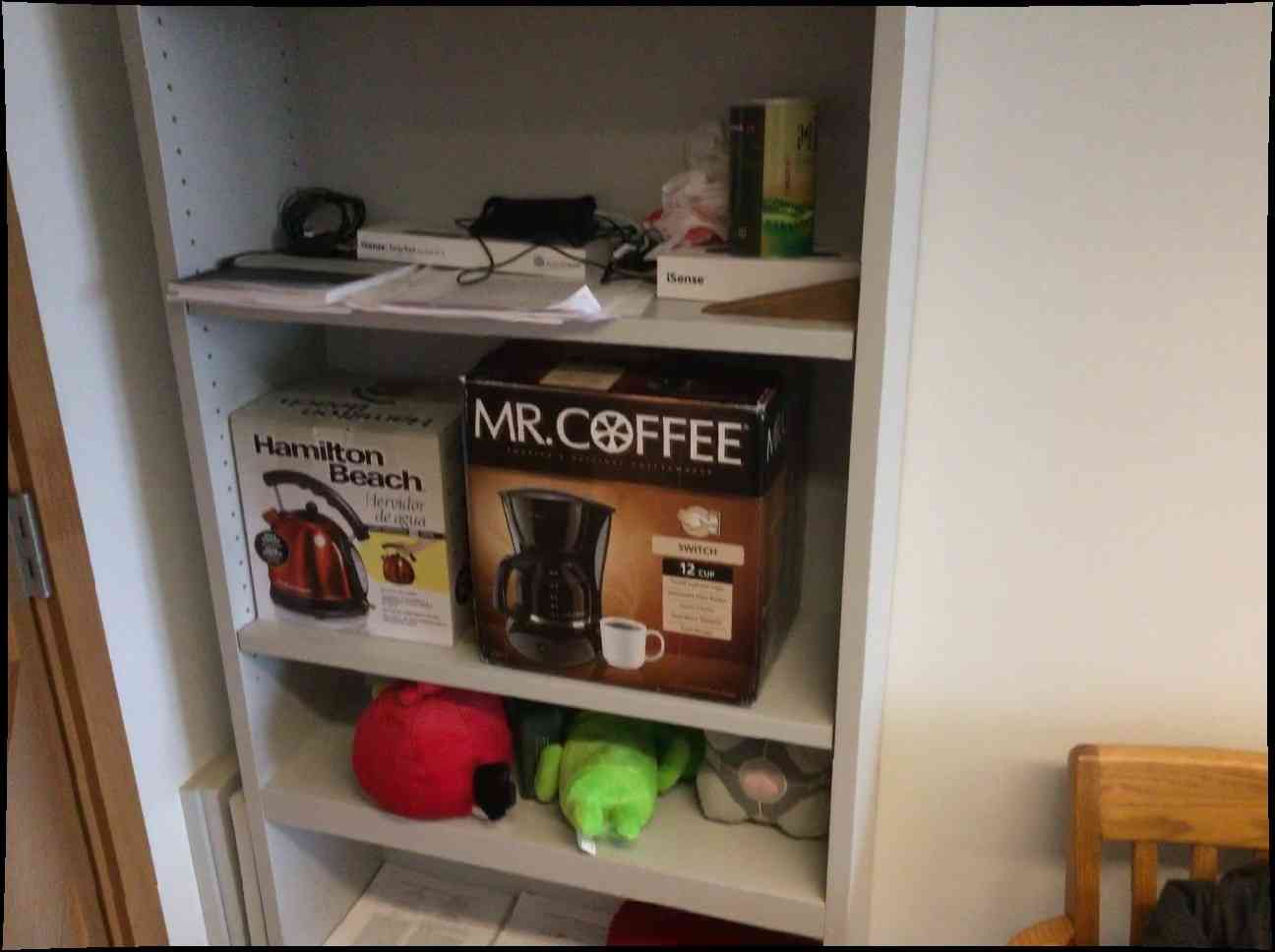}
    \captionof{figure}{In \autoref{sec_success_2}, The frame depicting the coffee machine named "Mr. Coffee" is successfully retrieved by frame\_localization.}
    \label{success_2}
\end{center}
}]

\begin{figure*}[h]
    \centering
    \begin{subfigure}[b]{0.49\textwidth}
        \centering
        \includegraphics[width=\linewidth]{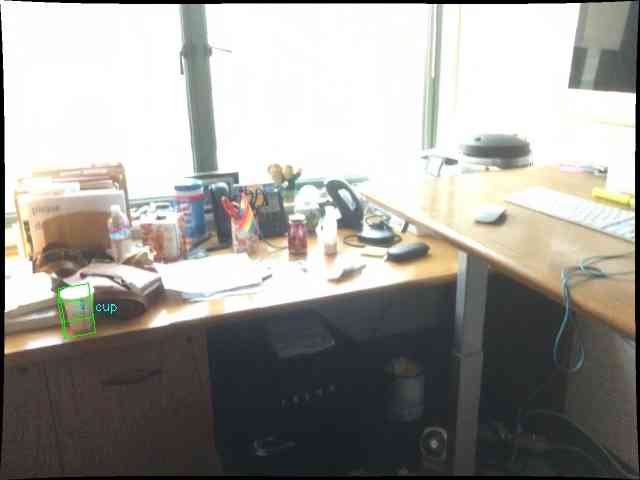}
        \caption{Object ID 2, category "cup".}
    \end{subfigure}
    \begin{subfigure}[b]{0.49\textwidth}
        \centering
        \includegraphics[width=\linewidth]{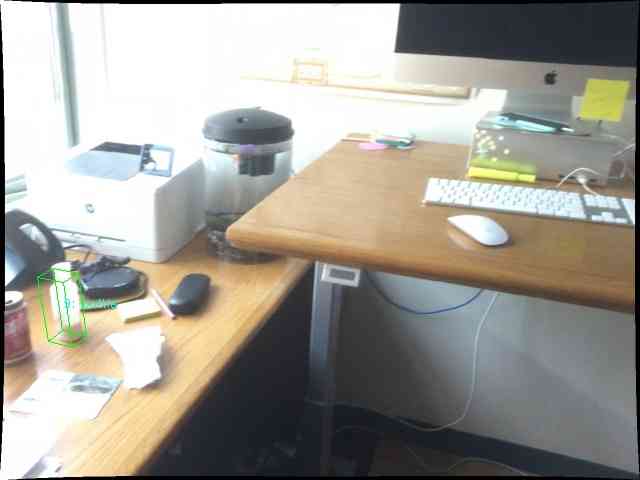}
        \caption{Object ID 9, category "bottle".}
    \end{subfigure}
    \vspace{1em}
    \begin{subfigure}[b]{0.49\textwidth}
        \centering
        \includegraphics[width=\linewidth]{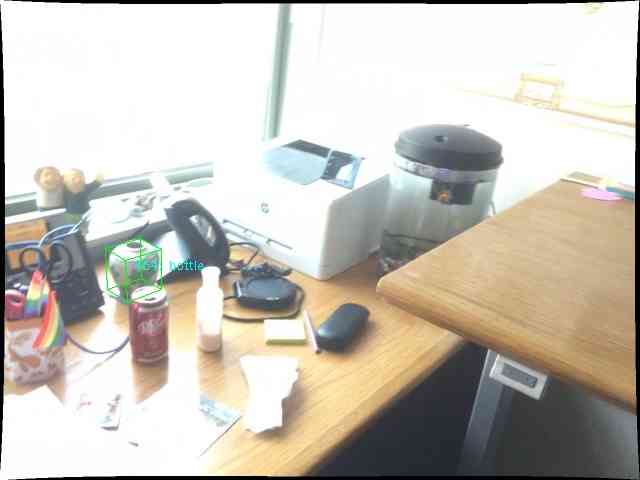}
        \caption{Object ID 454, category "bottle".}
    \end{subfigure}
    \begin{subfigure}[b]{0.49\textwidth}
        \centering
        \includegraphics[width=\linewidth]{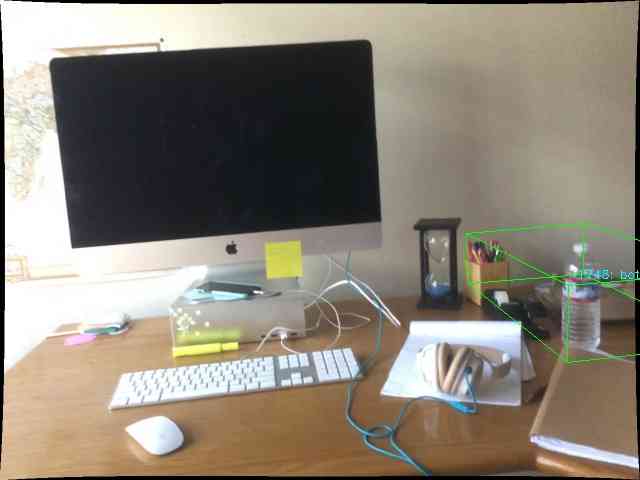}
        \caption{Object ID 1748, category "bottle".}
    \end{subfigure}
    \caption{The objects that are most relevant to "cup" are successfully retrieved by retrieve\_objects\_by\_appearance in \autoref{sec_success_3}.}
    \label{success_3}
\end{figure*}

\twocolumn[{%
\renewcommand\twocolumn[1][]{#1}%
\subsection{Example 3}
\label{sec_success_3}
\begin{itemize}
    \item \textbf{Question}: "Is there anything I can pour my soda into?"
    \item \textbf{Ground-Truth Answer}: "Yes, there is a cup to the left of the desk."
    \item \textbf{Predicted Answer}: "Yes, there is a cup and a bottle in the scene that you can pour your soda into."
\end{itemize}

This question is about functional reasoning. \method first uses retrieve\_objects\_by\_appearance to retrieve the objects that are most visually relevant to the phrase "cup" from the object memory. The captions of these objects are also returned by this tool via VLM captioning on the frames containing these objects, with 3D bounding boxes being the visual prompts, as shown in \autoref{success_3}. Then based on the retrieved objects, \method performs object\_VQA, which answers questions based on the images in \autoref{success_3}. By examining object ID 2 and object ID 454, \method becomes certain that there exists a container that the user can pour soda into. The CoT is provided as follows.

\begin{tcolorbox}[breakable]
Entering new AgentExecutor chain...

\textbf{Thought}: To determine if there is anything in the scene that can be used to pour soda into, I should look for objects that are typically used as containers for liquids. Common objects for this purpose include cups, bottles, and containers. I will start by retrieving objects that match these categories.

\textbf{Action}: retrieve\_objects\_by\_appearance

\textbf{Action Input}: "cup"

\textbf{Result}: The objects that satisfy 'cup' are {2: 'The object in the bounding box is a green cup or mug located on the left side of the table.', 454: 'The object in the bounding box is a bottle, specifically labeled "454 bottle". There is a green box around the highlighted area, and the object appears to have a label with green and black colors.', 1748: 'The object in the bounding box is a rectangular, black desk organizer positioned next to a black desk lamp on the desk. The organizer has compartments filled with pens, pencils, markers, and other office supplies.', 9: 'The object within the bounding box in the image is labeled as a "water bottle."'}

\textbf{Thought}: I have found some objects that could potentially be used to pour soda into, such as a cup and a bottle. I will now verify if these objects are suitable for pouring soda by asking specific questions about them.

\textbf{Action}: object\_VQA

\textbf{Action Input}: ("Is this object suitable for pouring soda into?", 2)

\textbf{Result}: Description: The object in the bounding box is a cup.

 Answer to the question: Although the object is not explicitly a cup, it resembles a cup and is likely suitable for pouring soda into.
 
 \textbf{Thought}: The object with ID 2 resembles a cup and is likely suitable for pouring soda into. I will now check the other objects to see if they are also suitable.

 \textbf{Action}: object\_VQA

 \textbf{Action Input}: ("Is this object suitable for pouring soda into?", 454)
 
 \textbf{Result}: Description: The object in the bounding box is a beverage can.

 Answer to the question: Yes, the object in question is suitable for pouring soda into, as it is shown as a bottle.
 
 \textbf{Thought}: The object with ID 454 is a bottle and is suitable for pouring soda into. I have confirmed that both the cup (ID 2) and the bottle (ID 454) are suitable for pouring soda into. 
 
 \textbf{Final Answer}: Yes, there is a cup and a bottle in the scene that you can pour your soda into.

Finished chain.
\end{tcolorbox}
}]

\clearpage
\section{Embodied Interactions}
In AI-Habitat simulator~\cite{savva2019habitat}, we equip \method with the following embodied actions:
\begin{itemize}
    \item \toolchat\texttt{CHAT}: Communicate with the user.
    \item \toolsearch\texttt{SEARCH}: Search for the target object by navigating in the apartment. We use Frontier-Based Exploration (FBE) as the navigation strategy.
    \item \toolgoto\texttt{GOTO}: Go to the target receptacle or object and look at it. We use A-star Algorithm for GOTO action.  
    \item \toolpick\texttt{PICK}: Pick an object in view. It is simplified as making the object disappear and storing the object ID as the inventory object.
    \item \toolplace\texttt{PLACE}: Place the inventory object in/on a receptacle in view. The Place Action will first examine the precondition for the placement by checking the bounding boxes of the inventory object and the receptacle and the relation "in" or "on".
    \item \toolopen\texttt{OPEN}: Open an articulated receptacle in view. Simplified as applying force to the joints of the articulated receptacles.
    \item \toolclose\texttt{CLOSE}: Close an articulated receptacle in view. Simplified as applying reversed force to the joints of the articulated receptacles.
\end{itemize}

\subsection{Two-Agent Pipeline}
We adopt the scenes from Habitat HSSD scene dataset\cite{hssd} for embodied tasks. We choose $118$ scenes from HSSD, replacing some rigid receptacles in the original scenes with articulated assets (fridge, microwave, etc) to enable OPEN and CLOSE actions. 

For each scene, $20$ different object layouts are created. In each layout, objects from various categories are placed on/into the receptacles in the scene using a unique object initialization algorithm, which initializes the positions of the objects according to their functionality (e.g. eggs and tomatoes are prioritized to be placed in the fridge rather than on the bed).

The embodied interaction episodes are generated based on two LLM agents: the User Agent (task designer) and the Assistant Agent (\method). The prompts for the user agent and the assistant agent are provided below. For \method, it is equipped with both the embodied actions and the perception tools.

\begin{tcolorbox}[colback=blue!10, colframe=blue!50!black, coltext=black, breakable]
You are a task designer interacting with a robot in a room. The room contains the following objects: \{\textbf{object\_list}\} and the following receptacles: \{\textbf{recep\_list}\}. Your goal is to engage in a casual conversation with the robot and assign it an open-ended task based on your needs.
~\\

Guidelines:

1. The task should involve no more than 2 objects from the room.

2. The robot should complete the task using basic actions like GOTO, OPEN, CLOSE, PICK, and PLACE.

3. If the robot asks for the location of an object, prompt it to search rather than giving explicit details.

4. Use general object categories instead of specific IDs (e.g., say "a dish sponge" instead of "dish sponge 1").

5. Adjust the task if the robot encounters difficulties.

Once the task is completed, express satisfaction and thank the robot.

Start by initiating a casual conversation and assigning a simple task!
\end{tcolorbox}

\begin{tcolorbox}[colback=red!10, colframe=red!50!black, coltext=black, breakable]
You are acting as a robot in an apartment. The available receptacles are:
\{\textbf{receptacles}\}
~\\

Your goal is to complete the task assigned by the user, with the following conditions:
~\\

Tools and Constraints:

You have one inventory slot, so you can carry only one object at a time.

You can use the following tools:

\{\textbf{tools}\}

ATTENTION:

1. Use the CHAT tool frequently to communicate in a casual manner, keeping the user informed of your progress.

2. For every action involving an object or receptacle, first GOTO the target and then perform actions like PICK, PLACE, OPEN, or CLOSE. Example: GOTO('glass'), then PICK('glass'); GOTO('fridge'), then OPEN('fridge').

3. Ensure your inventory is empty before picking up a new object.

4. The SEARCH tool can find objects by navigating the room, but it cannot check inside articulated receptacles (like fridges or microwaves). Use GOTO, OPEN, and CLOSE to check inside these receptacles.

5. Before completing the task, use CHAT to confirm the user’s satisfaction.
~\\

Use the following format:
~\\

Task: the initial task assigned by the user

Thought: you should always think about what to do

Action: the action to take, should be one of [\{\textbf{tool\_names}\}]

Action Input: the input to the action

Observation: the result of the action... (this Thought/Action/Action Input/Observation can repeat N times)

Final Answer: the chat message sent to user when the user is satisfied
~\\

Begin!

Task: \{\textbf{input}\}

Thought: \{\textbf{agent\_scratchpad}\}
\end{tcolorbox}

\clearpage
\twocolumn[{%
\renewcommand\twocolumn[1][]{#1}%
\subsection{Example 1}
\label{interaction_example_1}
\autoref{supp_interaction_1} shows an interaction example using the two-agent pipeline. Given the partial scene knowledge, the user agent asks the assistant agent (\method) to find two objects: a glass and a hard drive, to compare their surface reflection. \method then performs the SEARCH action, which will start Frontier-Based Exploration (FBE) until the target object is found in the view. During exploration, a glass is found on Table\_2, and \method reports this progress to the user agent. The user agent hints that the next object, the hard drive, is possibly located in an office. \method then uses QUERY\_DB tool and successfully retrieves the hard drive discovered by FBE during searching for the glass. \method then goes to the hard drive, picks it up, and places it on Tables\_2 where the glass is located for comparison, and finally accomplishes the task assigned by the user agent.
\begin{center}
    \includegraphics[width=\linewidth]{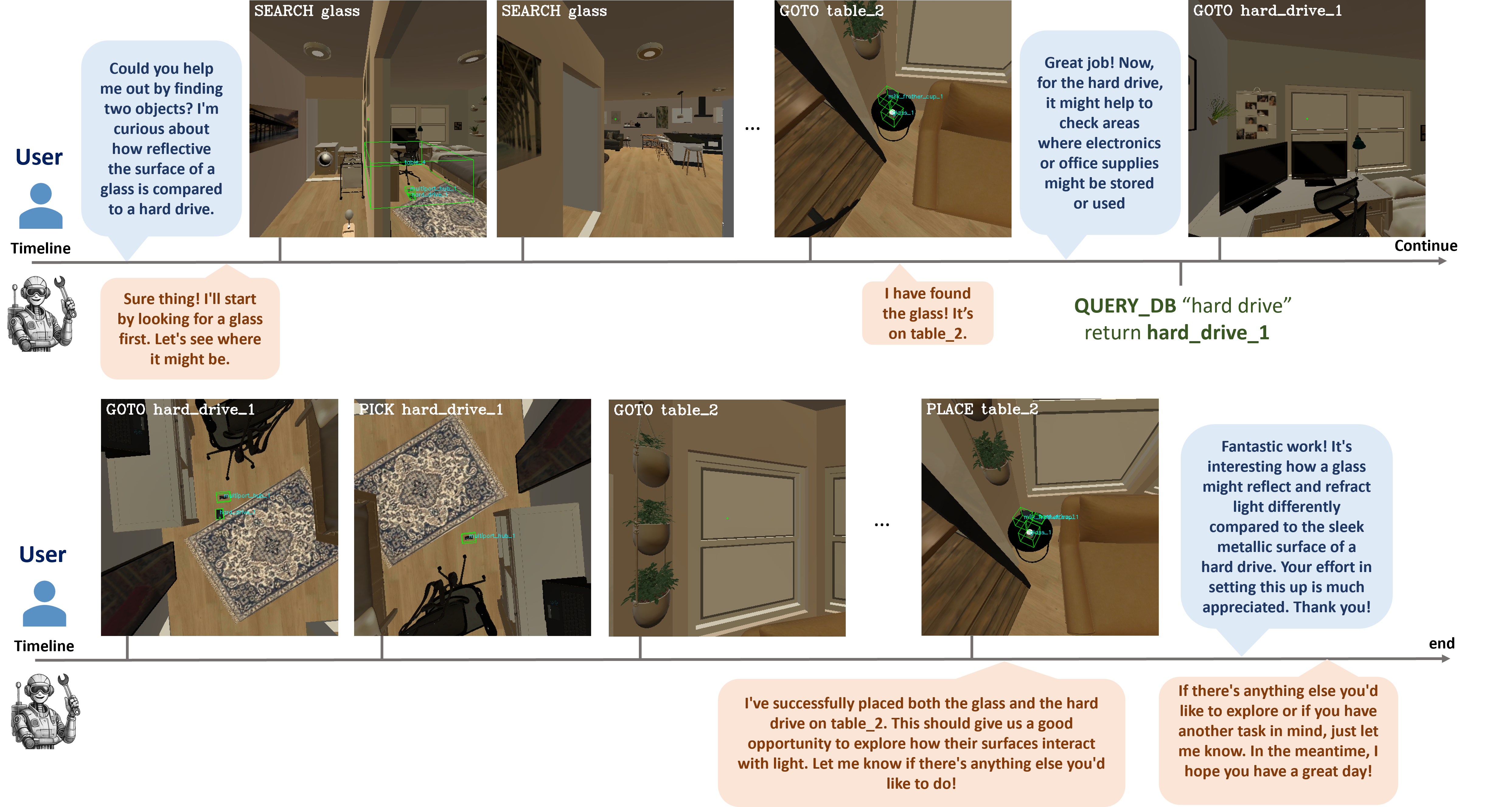}
    \captionof{figure}{An example of interaction data, which is detailed in \autoref{interaction_example_1}. \method finds the two objects (a glass and a hard drive) requested by the user agent and places them on the same table for comparison.}
    \label{supp_interaction_1}
\end{center}
}]

\clearpage
\twocolumn[{%
\renewcommand\twocolumn[1][]{#1}%
\subsection{Example 2}
In \autoref{supp_interaction_2}, the user agent requests \method to find a candy bar. After navigating through the entire apartment and checking the closed receptacles such as the fridge, \method still cannot find the candy bar, and report this issue to the user agent. The user agent then adjusts the task, asking \method to place a lamp on one of the tables. \method successfully retrieves the lamp stored in the object memory, which is discovered during searching for the candy. \method finally completes the adjusted tasks by picking up the lamp, navigating to a table, and placing the lamp on the table. This case shows that the user agent can flexibly change the task when it is too hard to be accomplished by the robot.
\begin{center}
    \includegraphics[width=\linewidth]{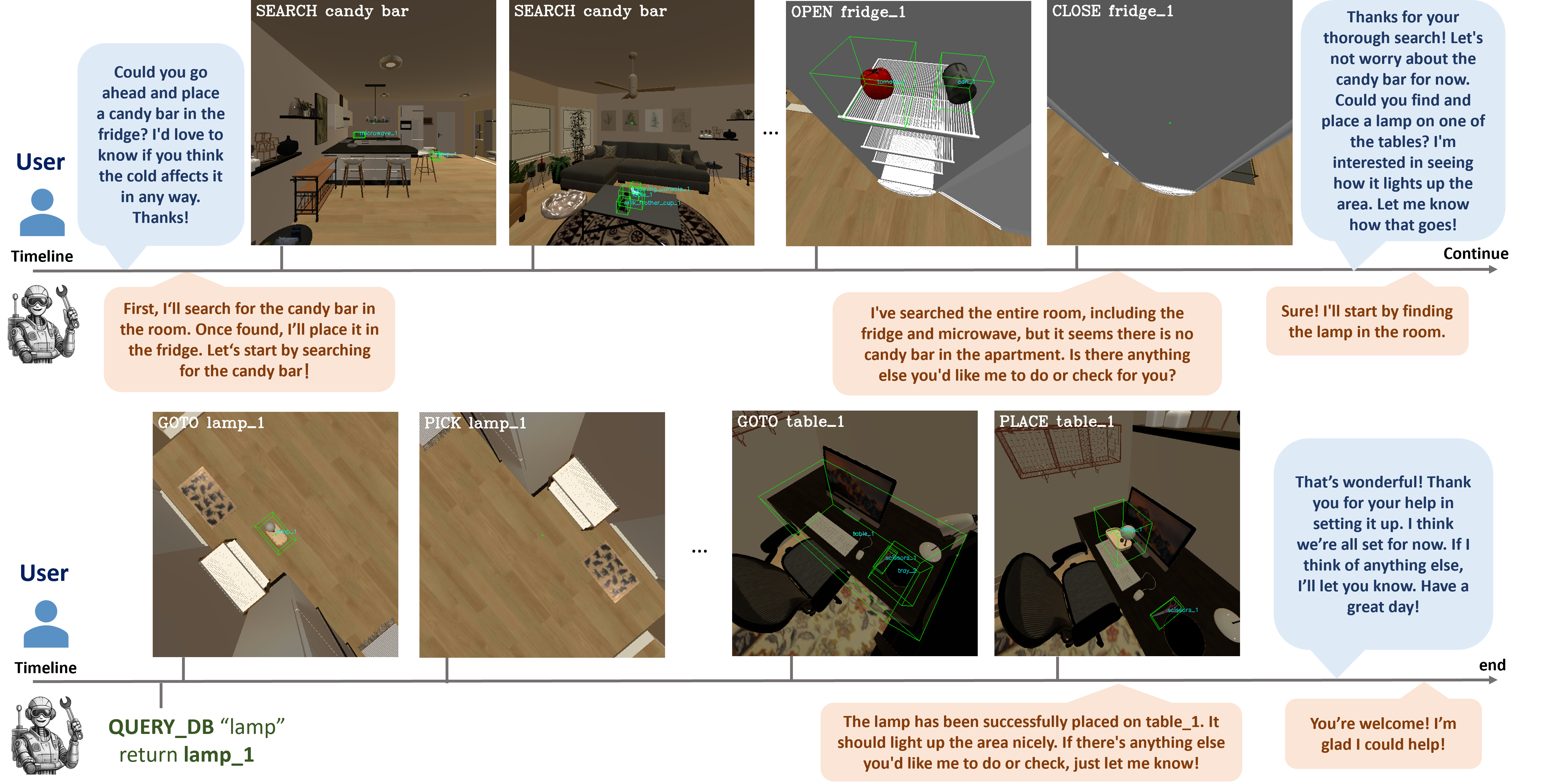}
    \captionof{figure}{When \method fails to accomplish the initial task assigned by the user (in this case, \method cannot find the candy bar after searching the entire apartment), the user agent will interactively adjust the task. Finally, \method completes the adjusted task by placing the lamb on a table.}
    \label{supp_interaction_2}
\end{center}
}]

\clearpage
\twocolumn[{%
\renewcommand\twocolumn[1][]{#1}%
\subsection{Example 3}
\autoref{supp_interaction_3} is an example that demonstrates the thought of \method during the episode. The user agent first initializes the task as setting up a cozy corner by placing a cushion on a table. \method finds the cushion after a simple search and plans to place it on one of the tables retrieved from the memory. However, the embodied action "PLACE" returns an exception due to the limited area of the first three tables. The thought of \method shows that \method can dynamically change its plan according to the situation. Finally, \method finished the task by successfully placing the cushion on the fourth table.
\begin{center}
    \includegraphics[width=\linewidth]{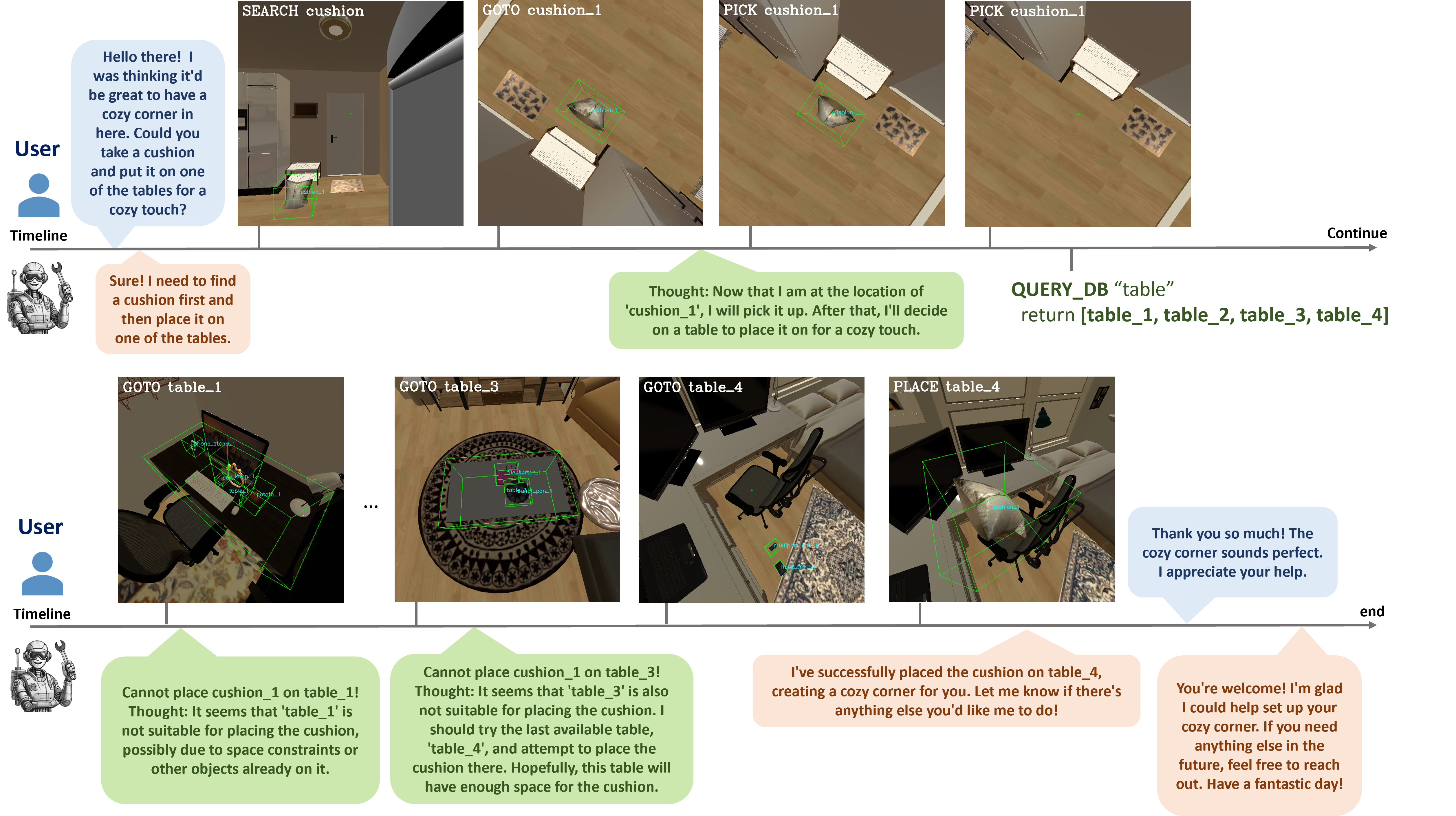}
    \captionof{figure}{In this case, \method is asked to set up a cozy corner using the cushion. After successfully finding the cushion, \method retrieves candidate tables from the memory for placement. \method fails to place the cushion on the first three tables due to spatial constraints. Finally, \method successfully places the cushion on the fourth table, demonstrating the planning ability of \method when experiencing failed embodied actions.}
    \label{supp_interaction_3}
\end{center}
}]

\clearpage
\section{Failure Cases}
In this section, we will provide some failure cases of \method on the perception tasks.

\begin{figure*}[t]
    \centering
    \begin{subfigure}[b]{0.33\textwidth}
        \centering
        \includegraphics[width=\linewidth]{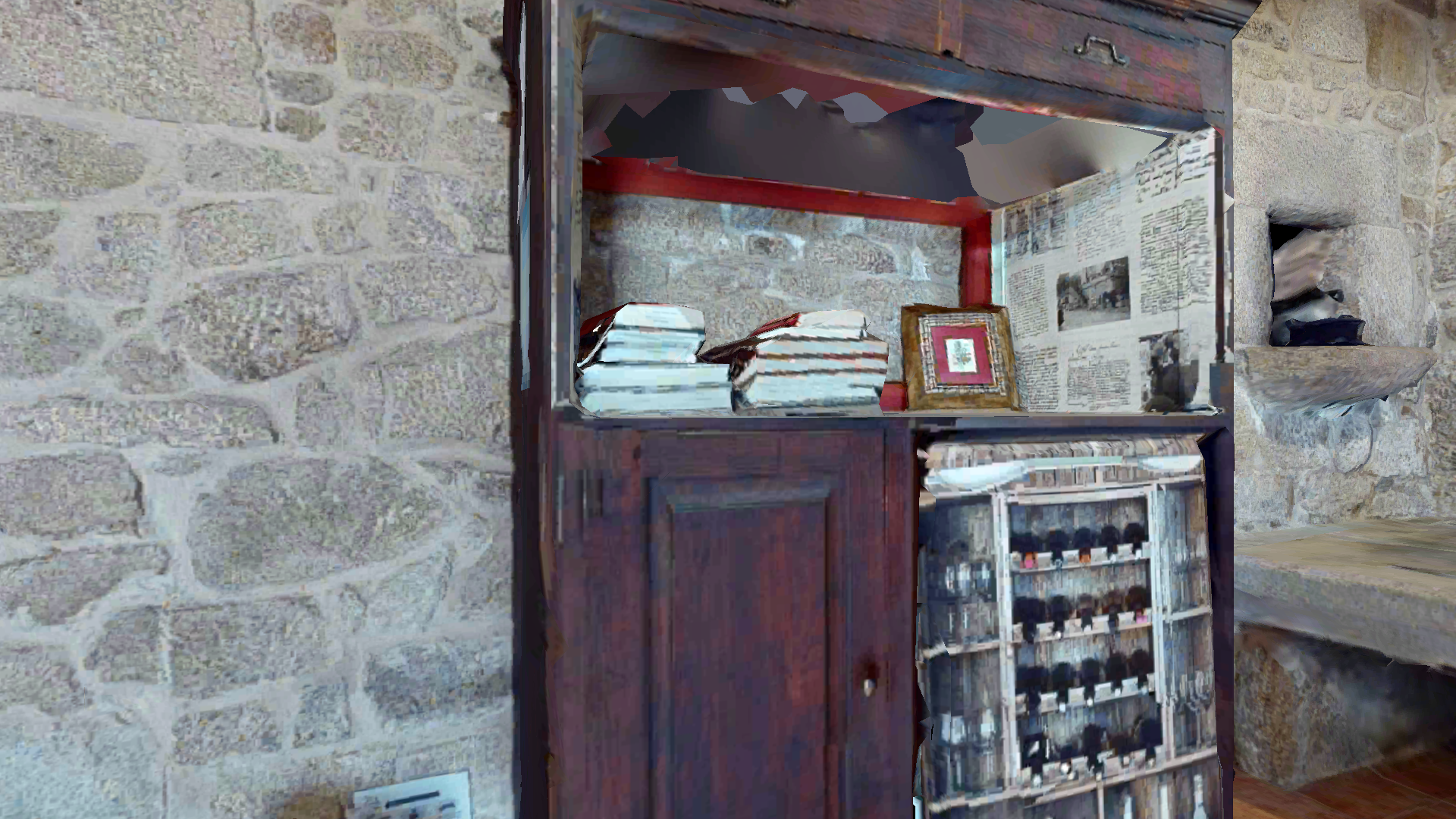}
        \caption{Frame for GT answer.}
        \label{fig:sub1}
    \end{subfigure}
    \begin{subfigure}[b]{0.33\textwidth}
        \centering
        \includegraphics[width=\linewidth]{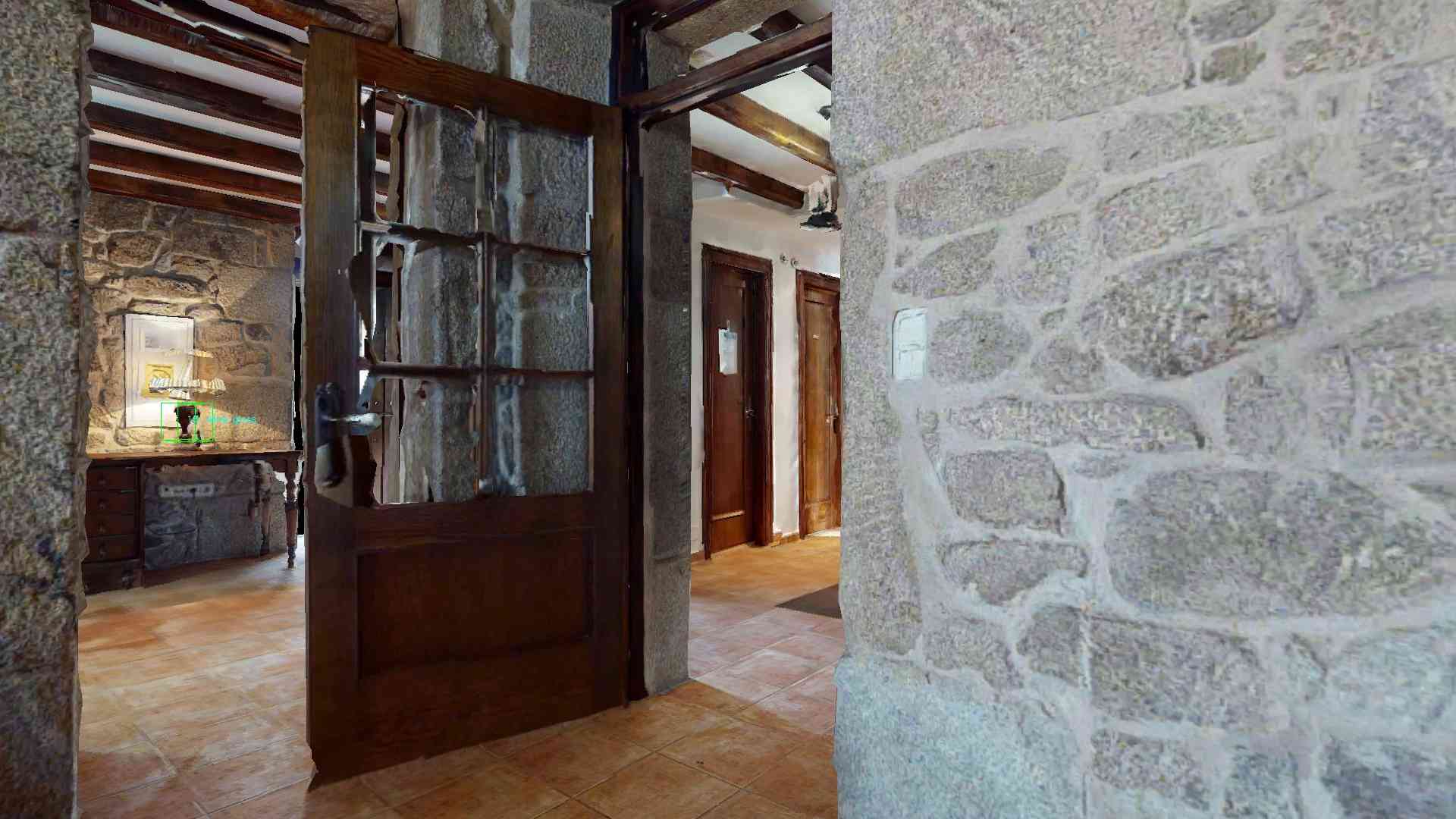}
        \caption{Frame 13 with false detection "wine glass".}
        \label{fig:sub2}
    \end{subfigure}
    \begin{subfigure}[b]{0.33\textwidth}
        \centering
        \includegraphics[width=\linewidth]{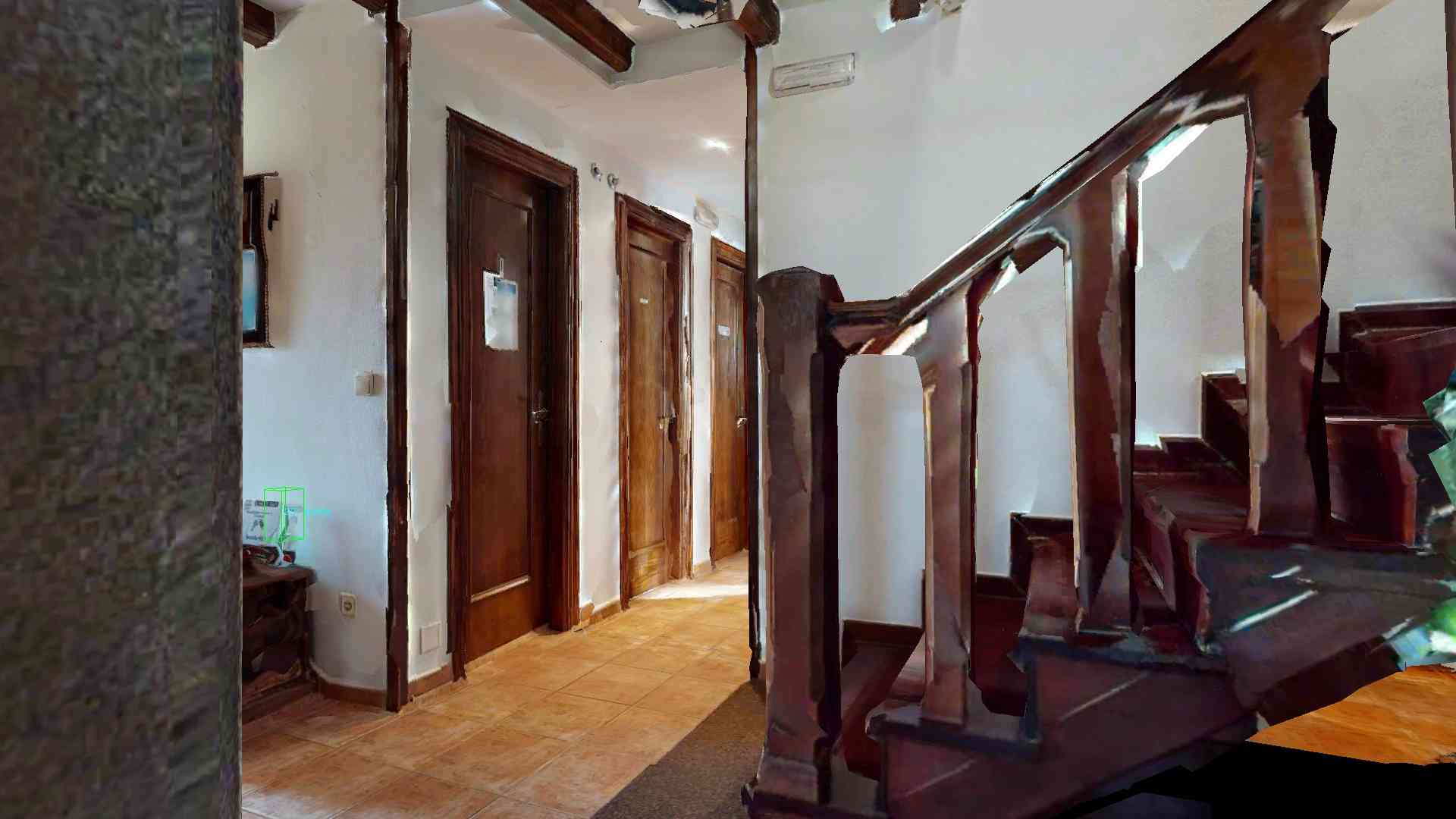}
        \caption{Frame 30 with false detection "bottle".}
        \label{fig:sub3}
    \end{subfigure}
    \caption{Due to poor image quality caused by scene reconstruction in HM3D, false object detections happen in frame 13 and frame 30. Inaccurate object detections finally cause wrong temporal localization.}
    \label{failure1}
\end{figure*}

\clearpage
\twocolumn[{%
\renewcommand\twocolumn[1][]{#1}%
\subsection{Failure Case 1}
\begin{itemize}
    \item \textbf{Question}: "Where can I find something to drink"
    \item \textbf{Ground-Truth Answer}: "In the fridge below the picture frame"
    \item \textbf{Predicted Answer}: "You can find something to drink in a cozy room within an older, rustic building, possibly a lodge or an inn, characterized by natural stone walls and floors, wooden beams, and wooden doors."
\end{itemize}
In this example, the object detector provides inaccurate object categories due to low image quality in HM3D datasets. A lamp in frame 13 is predicted as a wine glass and a box in frame 30 is predicted as a bottle, as shown in \autoref{failure1}. During the inference procedure, \method locates the keyframes based on the wrong categories of object 13 (wine glass) and object 22 (bottle), which finally leads to an inaccurate answer.

\begin{tcolorbox}[breakable]
Entering new AgentExecutor chain...

\textbf{Thought}: To find something to drink, I should look for objects related to drinking, such as 'wine glass' or 'bottle'. I will query the database to find these objects.

\textbf{Action}: query\_database

\textbf{Action Input}: "SELECT object\_id FROM Objects WHERE category IN ('wine glass', 'bottle')"

\textbf{Result}: [(13,), (22,)]

\textbf{Thought}: I have found object IDs 13 and 22, which are related to drinking. I need to determine in which frames these objects appear to locate them in the scene. I will query the database to find the frames for these objects.

\textbf{Action}: query\_database

\textbf{Action Input}: "SELECT frame\_id FROM Objects\_Frames WHERE object\_id IN (13, 22)"

\textbf{Result}: [(13,), (30,)]

\textbf{Thought}: I have found that the objects related to drinking appear in frames 13 and 30. I should now determine the environment or location of these frames to provide a more specific answer.

\textbf{Action}: frame\_VQA

\textbf{Action Input}: ("What is the environment or location of the frame?", 13)

\textbf{Result}: Description: This image depicts a cozy room that appears to be within an older, rustic building, perhaps a lodge or an inn. The room is characterized by a distinct blend of natural stone and wooden elements, which contribute to its warm and inviting ambiance.

Key elements in the image include:

1. **Natural Stone Walls \& Floor:** The walls and floor are made of natural stone, which adds to the rustic aesthetic. The stone floor is speckled with darker and lighter tones, creating a mosaic-like pattern.

2. **Wooden Beams:** There are exposed wooden beams supporting the ceiling, giving a sturdy and traditional feel.

3. **Wooden Doors:** Two wooden doors are present. The door on the left is ajar, revealing a room with a wooden table and a lamp stand. The door on the right stands tall against the stone wall.

4. **Lighting Fixtures:** A charming wall-mounted lamp with a cream shade is visible, providing warm lighting.

5. **Interior Details:** There is also a wooden chest with a unique metal clasp, suggesting it's either a storage piece or perhaps a decorative item.

6. **Wooden Cabinets:** The small alcove on the left features wooden cabinets, which implies functional storage space within the room.

7. **Entrance/Exit:** There is an open gate leading into the room on the left, furthering the visual appeal of the space, enhancing the sense of depth and accessibility within the interior.

The combination of natural materials, simple yet elegant furniture, combined with functional yet decorative items like the chest and the lamp, make this room not just living space, but a cozy retreat that invites warmth and comfort.

 Answer to the question: Based on the given image, the environment or location of the frame appears to be a residential or small-scale hotel setting, characterized by traditional and natural elements that include stone masonry, wood beams, and wooden floors and doors. The style suggests a countryside or historical district where such architecture is prominent, possibly indicating a location that values historical aesthetics and authenticity
 
 \textbf{Thought}: I have identified that the objects related to drinking (wine glass and bottle) are located in a cozy room within an older, rustic building, possibly a lodge or an inn. This environment is characterized by natural stone walls and floors, wooden beams, and wooden doors. Therefore, you can find something to drink in this cozy, rustic room setting.

\textbf{Final Answer}: You can find something to drink in a cozy room within an older, rustic building, possibly a lodge or an inn, characterized by natural stone walls and floors, wooden beams, and wooden doors.

Finished chain.
\end{tcolorbox}
}]

\clearpage
\twocolumn[{%
\renewcommand\twocolumn[1][]{#1}%
\subsection{Failure Case 2}

\begin{itemize}
    \item \textbf{Question}: "Could I put something on the chair by the front door?"
    \item \textbf{Ground-Truth Answer}: "Yes"
    \item \textbf{Predicted Answer}: "The chair by the front door has a small volume, so it might not be suitable for placing items on it."
\end{itemize}

In this example, \method successfully locates the chair (ID 107) by the front door as shown in \autoref{failure2}. Then, \method decides to query the volume of its 3D bounding box to answer the question "Could I put something on the chair by the front door". However, this chair is in the 2D texture of the scene, which fails 2D-3D lifting. The volume of its 3D bounding box (nearly being a plane) has a very small value, misleading \method into the wrong answer. In real-world applications, obtaining object depths through mirrors (windows, etc.) could have much noise, which should be taken into account for accurate perception. 

\begin{tcolorbox}[breakable]
Entering new AgentExecutor chain...
 
\textbf{Thought}: To answer this question, I need to determine if there is a chair by the front door and if it has enough space or surface area to place something on it. First, I will identify the objects categorized as "chair" and "door" in the scene. Then, I will check if any of these chairs are located near the front door.

\textbf{Action}: retrieve\_objects\_by\_environment

\textbf{Action Input}: "front door"

\textbf{Result}: The objects in the environment 'front door' are {493: 'door', 107: 'chair', 483: 'cubby shelf', 478: 'window', 482: 'photo frame'}

\textbf{Thought}: I have identified that there is a chair (object ID 107) near the front door. Now, I need to determine if this chair has enough space or surface area to place something on it. I will check the volume of the chair to get an idea of its size.

\textbf{Action}: query\_database

\textbf{Action Input}: "SELECT volume FROM Objects WHERE object\_id = 107"

\textbf{Result}: [(0.0006219363573111673,)]

\textbf{Thought}: The volume of the chair near the front door is quite small, suggesting it might not have a large surface area. However, without specific information about the chair's design or surface, it's difficult to definitively say whether something can be placed on it. Given the small volume, it is likely that the chair is not designed to hold items on its surface.

\textbf{Final Answer}: The chair by the front door has a small volume, so it might not be suitable for placing items on it.

Finished chain.
\end{tcolorbox}

\begin{center}
    \includegraphics[width=0.6\linewidth]{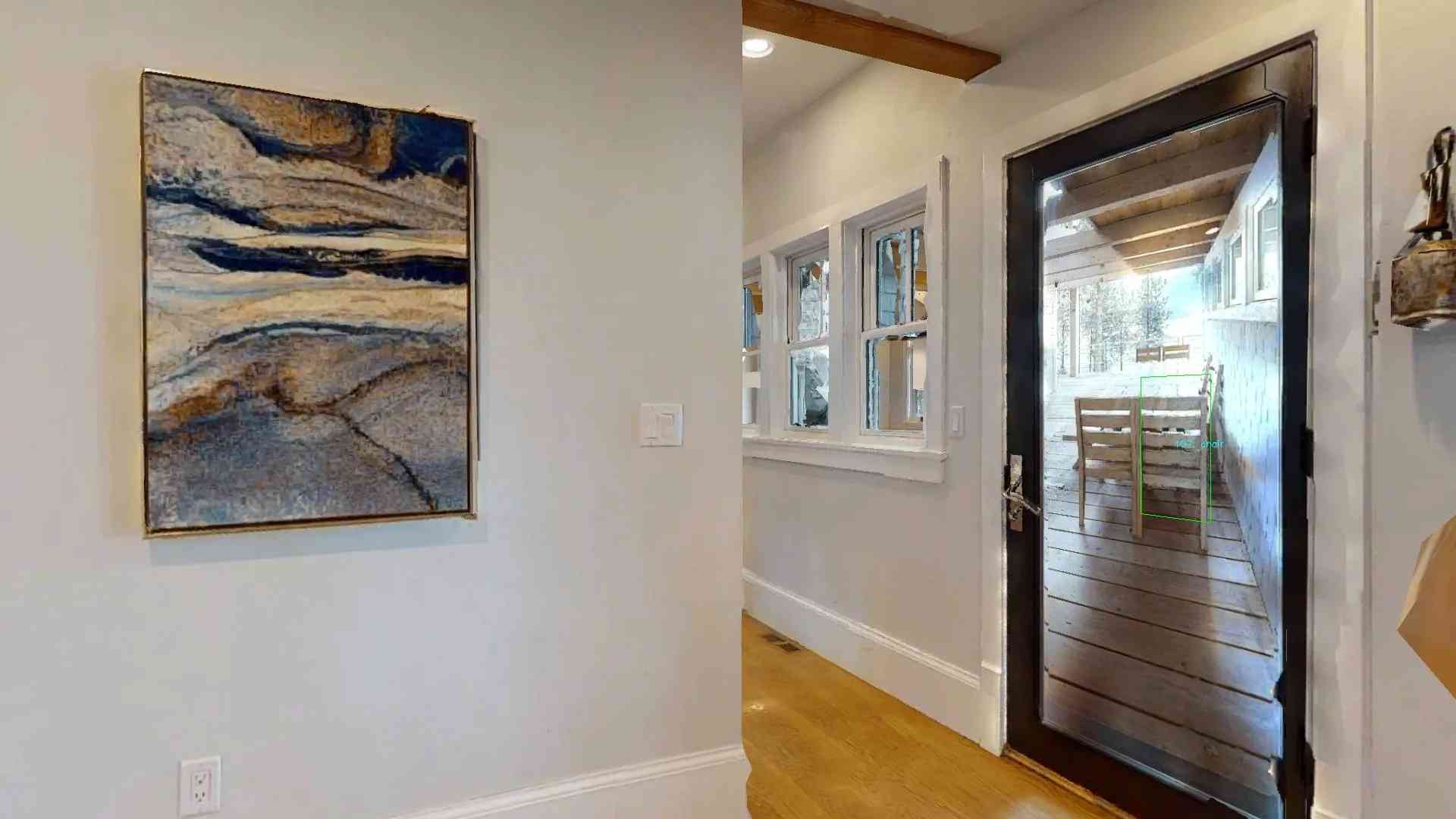}
    \captionof{figure}{The 2D bounding box of the detected chair is not successfully 3D-lifted since it is a 2D texture in the scene.}
    \label{failure2}
\end{center}
}]

\end{document}

%% file: configs.tex
\definecolor{Gray}{gray}{0.9}
\definecolor{mygreen}{rgb}{0.0, 0.5, 0.0}
\definecolor{myred}{rgb}{0.8, 0.25, 0.33}
\definecolor{myblue}{rgb}{0.19, 0.55, 0.91}
\definecolor{uclablue}{rgb}{0.15, 0.45, 0.68}
\definecolor{ucladblue}{rgb}{0.0, 0.33, 0.53}
\definecolor{ucladdblue}{rgb}{0.0, 0.23, 0.36}
\definecolor{uclagold}{rgb}{1.0, 0.82, 0.0}
\definecolor{ucladgold}{rgb}{1.0, 0.78, 0.17}
\definecolor{ucladdgold}{rgb}{1.0, 0.72, 0.11}
\definecolor{boxgreen}{rgb}{0.02, 0.66, 0.02}
\definecolor{boxred}{rgb}{0.66, 0.1, 0.1}
\definecolor{boxblue}{rgb}{0.01, 0.01, 0.73}

\usepackage{etoc}
\usepackage{lipsum}
\usepackage{wrapfig}
\usepackage{multicol}
\usepackage{mdframed}

\usepackage{mathtools}

\usepackage[utf8]{inputenc}   
\usepackage[T1]{fontenc}      
\usepackage{url}              
\usepackage{amsfonts}         
\usepackage{nicefrac}         
\usepackage{microtype}
\usepackage{graphicx}
\usepackage{amsmath,amssymb,mathbbol}
\usepackage{algorithmic}
\usepackage[linesnumbered,ruled,vlined]{algorithm2e}
\usepackage{acronym}
\usepackage{enumitem}
\usepackage{balance}
\usepackage{xspace}
\usepackage{setspace}
\usepackage[skip=3pt,font=small]{subcaption}
\usepackage[skip=3pt,font=small]{caption}
\usepackage[dvipsnames]{xcolor}
\usepackage{booktabs,tabularx,colortbl,multirow,array,makecell}
\usepackage{pifont}
\usepackage{xr-hyper}
\usepackage{tcolorbox}
\usepackage{adjustbox} 
\tcbuselibrary{breakable}

\usepackage{siunitx}   
\usepackage{soul}   
\usepackage{listings}
\definecolor{codegreen}{rgb}{0,0.6,0}
\definecolor{codegray}{rgb}{0.5,0.5,0.5}
\definecolor{codepurple}{rgb}{0.58,0,0.82}
\definecolor{backcolour}{rgb}{0.95,0.95,0.92}

\lstdefinestyle{mystyle}{
    backgroundcolor=\color{backcolour},   
    commentstyle=\color{codegreen},
    keywordstyle=\color{magenta},
    numberstyle=\tiny\color{codegray},
    stringstyle=\color{codepurple},
    basicstyle=\ttfamily\footnotesize,
    breakatwhitespace=false,         
    breaklines=true,                 
    captionpos=b,                    
    keepspaces=true,                 
    numbers=left,                    
    numbersep=5pt,                  
    showspaces=false,                
    showstringspaces=false,
    showtabs=false,                  
    tabsize=2
}

\makeatletter
\renewcommand{\paragraph}{%
  \@startsection{paragraph}{4}%
  {\z@}{0ex \@plus 0ex \@minus 0ex}{-1em}%
  {\hskip\parindent\normalfont\normalsize\bfseries}%
}
\makeatother

\definecolor{gblue}{HTML}{4285F4}
\definecolor{gred}{HTML}{DB4437}
\definecolor{ggreen}{HTML}{0F9D58}

\definecolor{mygray}{gray}{.92}

\newcommand{\supp}{\textit{Appendix}\xspace}

\acrodef{llm}[LLMs]{large language models}
\acrodef{vla}[VLA]{vision-language-action}
\acrodef{vl}[VL]{vision-language}
\acrodef{ocot}[O-CoT]{Object-centric Chain-of-Thought}
\acrodef{cot}[CoT]{Chain-of-Thought}
\acrodef{lvlm}[LVLM]{large vision-language models}

\newcommand{\method}{\textit{Embodied VideoAgent}\xspace}